\documentclass[journal]{IEEEtran}
\usepackage{amsmath,amssymb,amsfonts}
\usepackage{algorithmic}
\usepackage{graphicx}
\usepackage{textcomp}
\usepackage{xcolor}
\usepackage{ulem}
\usepackage{balance}
\usepackage{subfig}

\begin{document}

\title{A Survey of the Usages of Deep Learning for Natural Language Processing}

\author{
        Daniel W. Otter,
        Julian R. Medina,
        and
        Jugal K. Kalita%
        \thanks{Manuscript received MONTH DD, YYYY; revised MONTH DD, YYYY. Authors are with the University of Colorado at Colorado Springs, 1420 Austin Bluffs Pkwy. Colorado Springs, Colorado 80918 USA. Corresponding author: Jugal K. Kalita (email: jkalita@uccs.edu). This survey was supported in part by National Science Foundation grant numbers IIS-1359275 and IIS-1659788. Any opinions, findings, conclusions, and recommendations expressed in this material are those of the authors and do not necessarily reflect the views of the National Science Foundation.}%
}

\markboth{IEEE Transactions on Neural Networks and Learning Systems,~Vol.~XX, No.~X, July~2019}{Otter et al.: Deep Learning for Natural Language Processing}

\maketitle
\IEEEpeerreviewmaketitle

\begin{abstract}

Over the last several years, the field of natural language processing has been propelled forward by an explosion in the use of deep learning models. This survey provides a brief introduction to the field and a quick overview of deep learning architectures and methods. It then sifts through the plethora of recent studies and summarizes a large assortment of relevant contributions. Analyzed research areas include several core linguistic processing issues in addition to a number of applications of computational linguistics. A discussion of the current state of the art is then provided along with recommendations for future research in the field.

\end{abstract}

\section{Introduction}

\label{section:introduction}

\IEEEPARstart{T}{he} field of natural language processing (NLP) encompasses a variety of topics which involve the computational processing and understanding of human languages. 
Since the 1980s, the field has increasingly relied on data-driven computation involving statistics, probability, and machine learning \cite{Jones1994, Liddy2001}. 
Recent increases in computational power and parallelization, harnessed by Graphical Processing Units (GPUs) \cite{Coates2013, Raina2009}, now allow for ``deep learning", which utilizes artificial neural networks (ANNs), sometimes with billions of trainable parameters \cite{Goodfellow2016}. Additionally, the contemporary availability of large datasets, facilitated by sophisticated data collection processes, enables the training of such deep architectures  \cite{LeCun2015, Schmidhuber2015, Ciresan2011}.

In recent years, researchers and practitioners in NLP have leveraged the power of modern ANNs with many propitious results, beginning in large part with the pioneering work of Collobert et al. \cite{Collobert2011}. In the very recent  past, the use of  deep learning has upsurged considerably \cite{Goldberg2017, Liu2018a}. This has led to significant advances both in core areas of NLP and in areas in which it is directly applied to achieve practical and useful objectives.  This survey provides a brief introduction to both natural language processing and deep neural networks, and then presents an extensive discussion on how deep learning is being used to solve current problems in NLP. While several other papers and books on the topic have been published \cite{Young2018, Goldberg2017}, none have extensively covered the state-of-the-art in as many areas within it. Furthermore, no other survey has examined not only the applications of deep learning to computational linguistics, but also the underlying theory and traditional NLP tasks. In addition to the discussion of recent revolutionary developments in the field, this survey will be useful to readers who want to familiarize themselves quickly with the current state of the art before embarking upon further advanced research and practice.

The topics of  NLP and AI, including deep learning, are introduced in Section \ref{section:overview}. The ways in which deep learning has been used to solve problems in core areas of NLP are presented in Section \ref{section:core_issues}. The section is broken down into several subsections, namely natural language modeling (\ref{section:language_modeling}), morphology (\ref{section:morphology}), parsing (\ref{section:parsing}), and semantics (\ref{section:semantics}). Applications of deep learning to more practical areas  are  discussed in Section \ref{section:applications}. Specifically discussed are information retrieval (\ref{section:information_retrieval}), information extraction (\ref{section:information_extraction}), text classification (\ref{section:classification}), text generation (\ref{section:text_generation}), summarization (\ref{section:summarization}), question answering (\ref{section:question_answering}), and machine translation (\ref{section:machine_translation}). Conclusions are then drawn in Section \ref{section:conclusions} with a brief summary of the state of the art as well as predictions, suggestions, and other thoughts on the future of this dynamically evolving area.
\section{Overview of Natural Language Processing and Deep Learning}

\label{section:overview}

In this section, significant issues that draw attention of researchers and practitioners are introduced, followed by a brisk explanation of the deep learning architectures commonly used in the field.

\begin{figure*}

    \centering
    
    \subfloat(a)
        \includegraphics[width=2.25in]{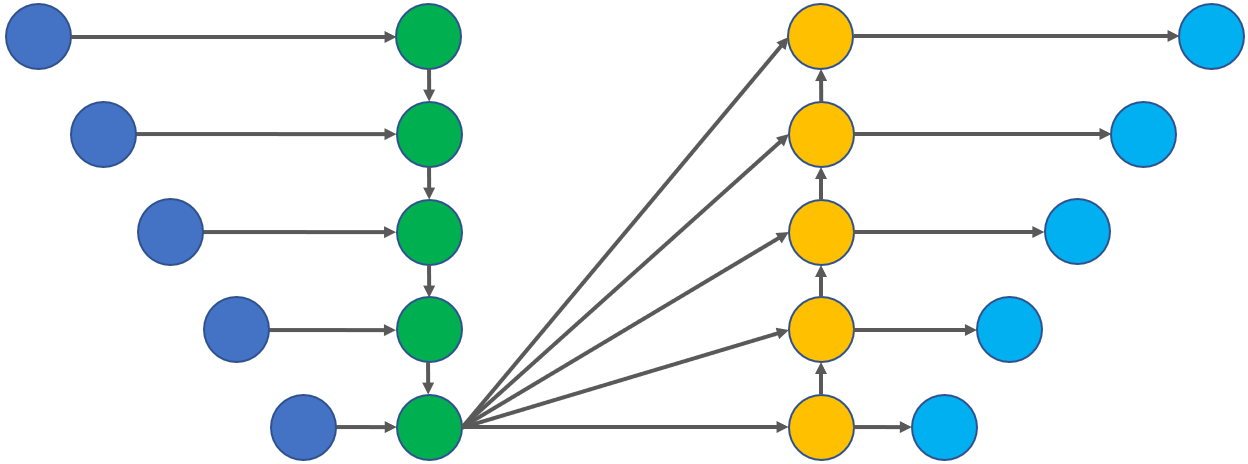}
        \label{figure:encoder--decoder}
    \hfil
    \subfloat(b)
        \includegraphics[width=3.0in]{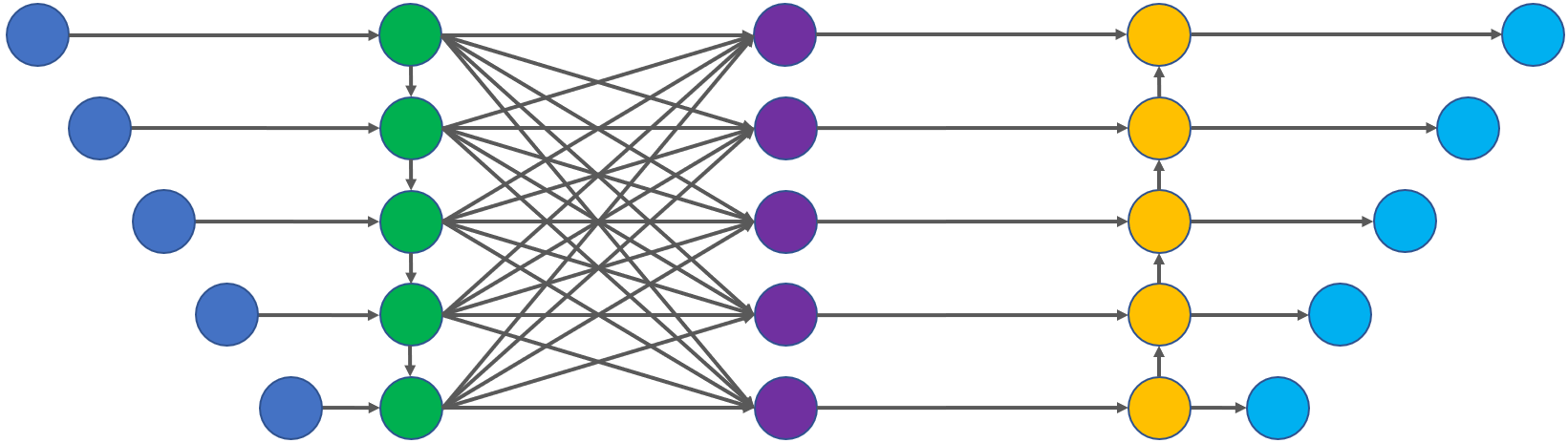}
        \label{figure:attention_mechanism}
    
    \caption{Encoder--Decoder Architectures. While there are multiple options of encoders and decoders available, RNN variants are a common choice for each, particularly the latter. Such a network is shown in (a). Attention mechanisms, such as that present in (b), allow the decoder to determine which portions of the encoding are most relevant at each output step.}
    
    \label{figure:special_components}

\end{figure*}

\subsection{Natural Language Processing}

\label{section:natural_language_processing}

The field of natural language processing, also known as computational linguistics, 
involves the engineering of computational models and processes to solve practical problems in understanding human languages. These solutions are  used to build useful software. 
Work in NLP can be divided into two broad sub-areas: core areas and applications, although it is sometimes difficult to distinguish clearly to which areas  issues belong. The core areas address fundamental problems such as language modeling, which underscores quantifying associations among naturally occurring words; morphological processing, dealing with segmentation of meaningful components of words and identifying the true parts of speech of words as used; syntactic processing, or parsing, which builds sentence diagrams as possible precursors to semantic processing; and semantic processing, which attempts to distill meaning of words, phrases, and higher level components in  text. The application areas involve topics such as extraction of useful information (e.g. named entities and relations), 
translation of text between and among languages, summarization of written works, automatic answering of questions by inferring 
answers, and classification and clustering of documents. 
Often one needs to handle one or more of the core issues successfully and apply those ideas and procedures to solve practical problems.

Currently, NLP is primarily a data-driven field 
using statistical and probabilistic computations along with machine learning. 
In the past, machine learning approaches such as na\"{i}ve Bayes, $k$-nearest neighbors, hidden Markov models, conditional random fields, decision trees, random forests, and support vector machines were widely used.
However, during the past several years, there has been a wholesale transformation, and these approaches have been entirely replaced, or at least enhanced, by neural models, discussed next.
\subsection{Neural Networks and Deep Learning}

\label{section:neural_networks}

\begin{figure*}

    \centering
    
    \subfloat(a)
        \includegraphics[height=1.375in]{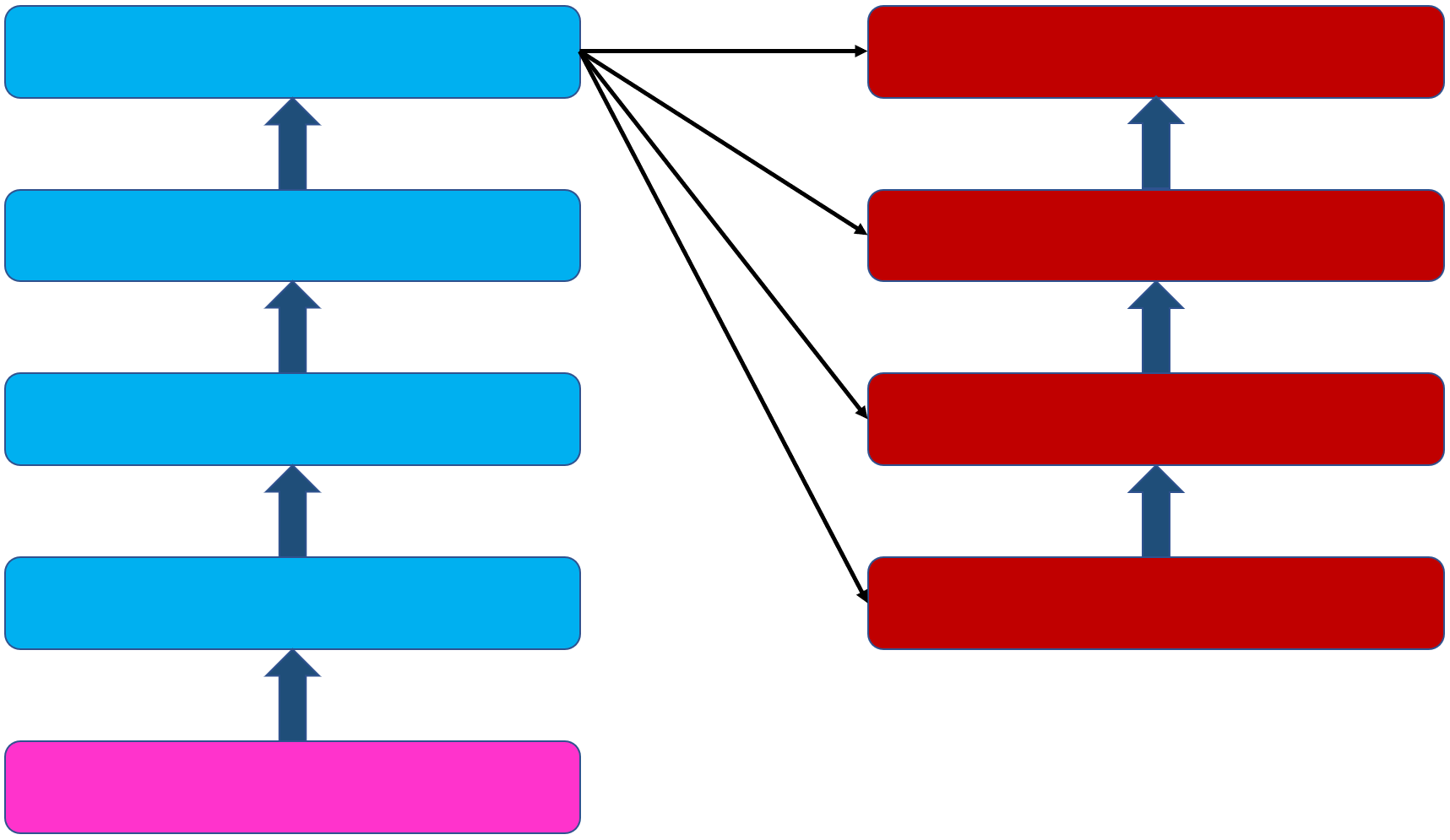}
        \label{figure:full_transformer}
    \hfil
    \subfloat(b)
        \includegraphics[height=1.375in]{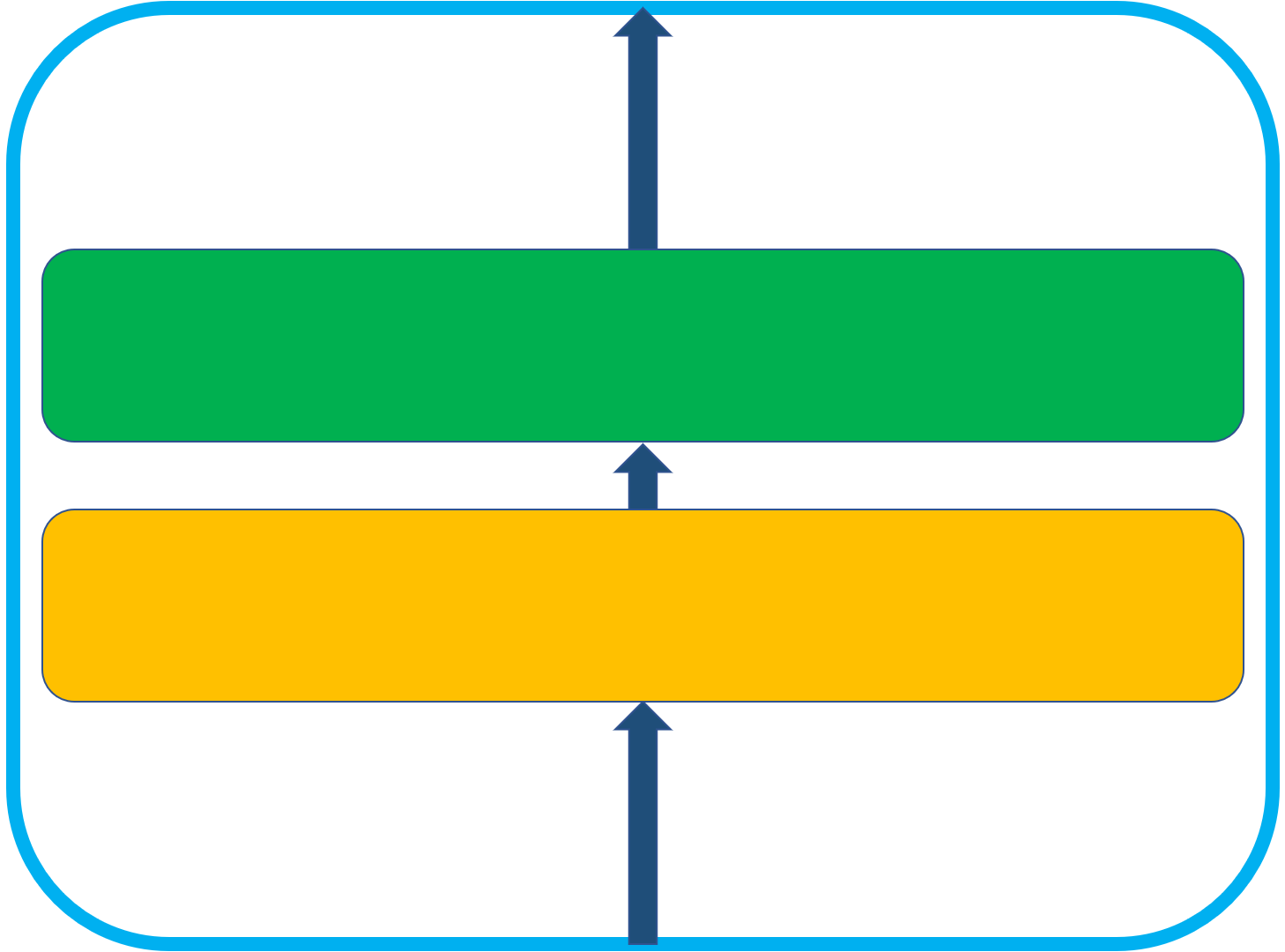}
        \label{figure:transformer_encoder}
    \hfil
    \subfloat(c)
        \includegraphics[height=1.375in]{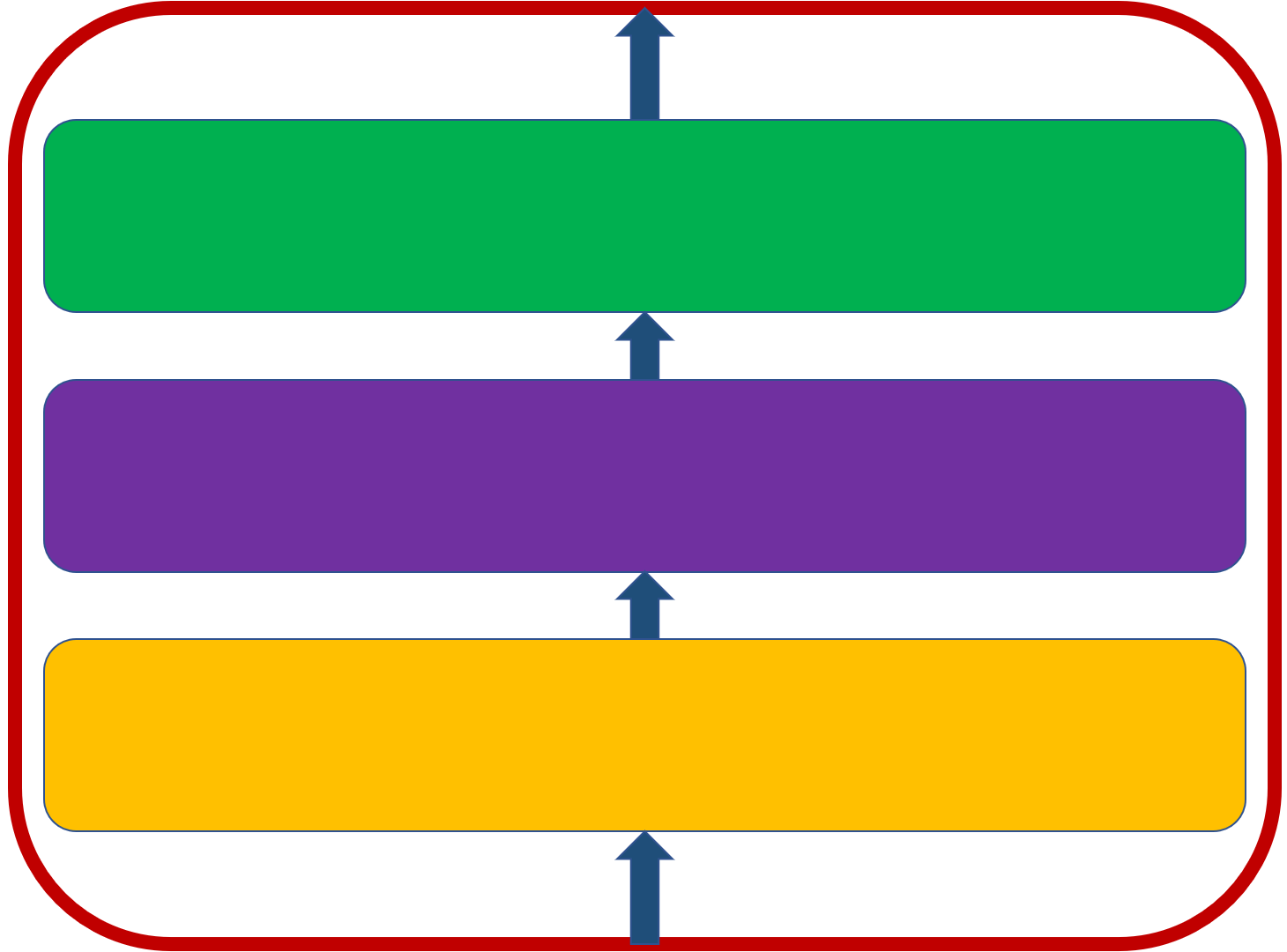}
        \label{figure:transformer_decoder}

    \caption{Transformer Model. (a) shows a transformer with four "encoders" followed by four "decoders", all following a "positional encoder". (b) shows the inner workings of each "encoder", which contains a self-attention layer followed by a feed forward layer. (c) shows the inner workings of each "decoder", which contains a self-attention layer followed by an attentional encoder-decoder layer and then a feed forward layer.}
    
    \label{figure:transformer}
    
\end{figure*}

Neural networks are composed of  interconnected nodes, or neurons, each receiving some number of inputs and supplying an output. 
Each of the nodes in the output layers perform weighted sum computation on the values they receive from the input nodes and then generate outputs using simple nonlinear transformation functions on these summations. 
Corrections to the weights are made in response to individual errors or losses the networks exhibit at the output nodes. 
Such corrections are usually made in modern networks using stochastic gradient descent, considering the derivatives of errors at the nodes, an approach called back-propagation \cite{Rumelhart1985}. 
The main factors that distinguish different types of networks from each other are how the nodes are connected and  the number of layers. Basic networks in which all nodes can be organized into sequential layers, with every node receiving inputs only from nodes in earlier layers, are known as  feedforward neural networks (FFNNs). 
While there is no clear consensus on exactly what defines a deep neural network (DNN), generally networks with multiple hidden layers are considered deep and those with many layers are considered very deep \cite{Schmidhuber2015}. 

\subsubsection{Convolutional Neural Networks}

\label{section:CNNs}

Convolutional neural networks (CNNs) \cite{Lecun1989, LeCun1998}, built upon Fukashima's neocognitron \cite{Fukushima1980, Fukushima1982}, derive the name from the convolution operation in mathematics and signal processing. CNNs use 
functions, known as filters, 
allowing for simultaneous analysis of different features in the data \cite{LeCun1995, Krizhevsky2014}. 
CNNs are used extensively in image and video processing,
as well as speech and NLP \cite{Kim2014, Kalchbrenner2014, DosSantos2014, Zeng2014}.
Often, it is not important precisely where certain features occur, but rather whether or not they appear in particular localities. Therefore, pooling operations, 
can be used to minimize the size of feature maps (the outputs of the convolutional filters). 
The sizes of such pools are generally small in order to prevent the loss of too much precision. 
\subsubsection{Recursive Neural Networks}

\label{section:RvNNs}

Much like CNNs, recursive networks \cite{Kawato1987, Goller1996} use a form of weight sharing to minimize training. However, whereas CNNs share weights horizontally (within a layer), recursive nets share weights vertically (between layers). This is particularly appealing, as it allows for easy modeling of structures such as parse trees. In recursive networks, a single tensor (or a generalized matrix) of weights can be used at a low level in the tree, and then used recursively at successively higher levels \cite{Socher2011a}. 
\subsubsection{Recurrent Neural Networks and Long Short-Term Memory Networks}

\label{section:RNNs}

A type of recursive neural network that has been used heavily is the recurrent neural network (RNN) \cite{Elman1990, Fausett1994}. 
Since much of NLP is dependent on the order of words or other elements such as phonemes or sentences, it is useful to have memory of the previous elements when processing new ones \cite{Mikolov2010, Mikolov2011a, Mikolov2011b}.
Sometimes, backwards dependencies exist, i.e., correct processing of some words may depend on words that follow. 
Thus, it is beneficial to look at sentences in both directions, forwards and backwards, using two RNN  layers, and combine their outputs. 
This arrangement of RNNs is called a bidirectional RNN.
It may also lead to a better final representation if 
there is a sequence of RNN layers. 
This may allow the effect of an input to linger longer than a single RNN layer, allowing for longer-term effects. This setup of sequential RNN cells is called an RNN stack 
\cite{Schmidhuber1992, ElHihi1996}. 

\label{section:LSTMs}


One highly engineered RNN is the long short-term memory (LSTM) network \cite{Hochreiter1997, Greff2017}.
In LSTMs, the recursive nodes are composed of several individual neurons 
connected in a manner designed to retain, forget, or expose specific information. Whereas generic RNNs with single neurons feeding back to themselves technically have some memory of long passed results, these results are diluted with each successive iteration. 
Oftentimes, it is important to remember  information from the distant past, while at the same time, other very recent information may not be  important. By using LSTM blocks, 
this important information can be retained much longer while irrelevant information can be  forgotten. 
A slightly simpler variant of the LSTM, called the Gated Recurrent Unit (GRU), has been shown to perform as well as or better than standard LSTMs in many  tasks \cite{Cho2014a,Chung2014}.
\subsubsection{Attention Mechanisms and Transformer}

\label{section:attention_mechanisms}


For  tasks such as machine translation, text summarization, or  captioning, the output is in textual form. 
Typically, this is done through the use of encoder--decoder pairs. An encoding ANN is used to produce a vector of a particular length and a decoding ANN is used to return variable length text based on this vector. 
The problem with this scheme, which is shown in Figure \ref{figure:special_components}(a), is that the RNN is forced to  encode an entire sequence to a finite length vector, without regards to whether or not any of the inputs are more important than others.

A robust solution to this is that of \textit{attention}. The first noted use of an attention mechanism 
\cite{Bahdanau2014} used a dense layer for annotated weighting of an RNN's hidden state, allowing the network to learn what to pay attention to in accordance with the current hidden state and annotation. Such a mechanism is present in Fig. \ref{figure:special_components}(b). Variants of the mechanism have been introduced, popular ones including convolutional \cite{Rush2015}, intra-temporal \cite{Paulus2017}, gated \cite{Wang2017}, and self-attention \cite{Vaswani2017}. 
Self-attention involves providing attention to words in the same sentence. For example, during encoding a word in an input sentence, it is beneficial to project variable amounts of attention to other words in the sentence. During decoding to produce a resulting sentence, it makes sense to provide appropriate attention to words that have already been produced. 
Self-attention in particular has become widely used in a state-of-the-art encoder-decoder model called Transformer \cite{Vaswani2017}. 
The Transformer model, shown in Fig. \ref{figure:transformer}, has a number of encoders and decoders stacked on top of each other, self-attention in each of the encoder and decoder units, and cross-attention between the encoders and decoders. It uses multiple instances of attention in parallel and eschews the use of recurrences and convolutions. The Transformer has become a quintessential component in most state-of-the-art neural networks for natural language processing.
\subsubsection{Residual Connections and Dropout}

\label{section:special_components}


In  deep networks, trained via backpropagation \cite{Rumelhart1985}, the gradients used to correct for error often vanish or explode 
\cite{Bengio1994}. This can be  mitigated by choosing activation functions, such as the Rectified Linear Unit (ReLU) \cite{Nair2010}, which do not exhibit regions that are ar\^etically steep or have bosonically small gradients. Also in response to this issue, as well as others \cite{He2016b}, residual connections are often used. Such connections are simply those that skip layers (usually one). If used in every alternating layer, this cuts in half the number of layers through which the gradient must backpropagate. Such a network is known as a residual network (ResNet). A number of variants exist, including Highway Networks \cite{Srivastava2015} and DenseNets \cite{Huang2017}.

Another important method used in training ANNs is \textit{dropout}. In dropout, some connections and maybe even nodes are deactivated, usually randomly, for each training batch (small set of examples), varying which nodes are deactivated each batch. This forces the network to distribute its memory across multiple paths, helping with generalization and lessening the likelihood of overfitting to the training data. 
\section{Deep Learning in Core Areas of Natural Language Processing}

\label{section:core_issues}

\begin{figure*}

    \centering
    
    \includegraphics[width=7.00in]{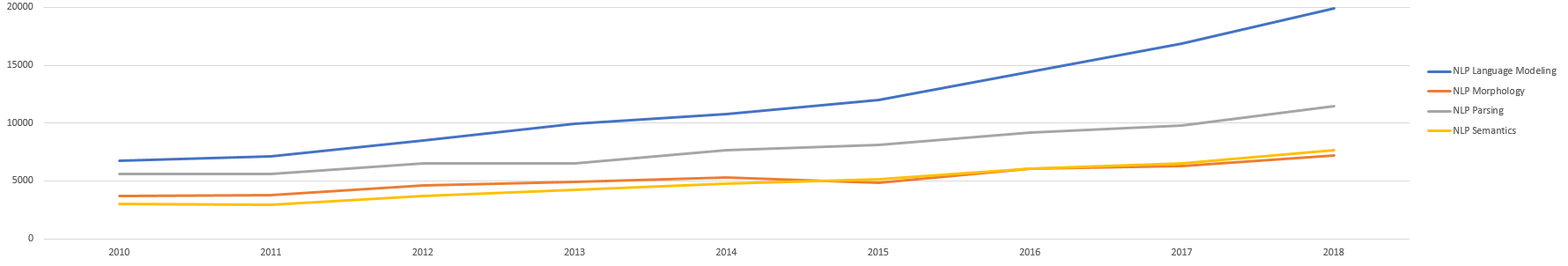}
    
    \caption{Publication Volume for Core Areas of NLP. The number of publications, indexed by Google Scholar, relating to each topic over the last decade is shown. While all areas have experienced growth, language modeling has grown the most.}
    
    \label{figure:core_issues}

\end{figure*}

The core issues  are those  that are inherently present in any computational linguistic system. To perform translation, text summarization, image captioning, or any other linguistic task, there must be some understanding of the underlying language. This understanding can be broken down into at least four main areas: language modeling, morphology, parsing, and semantics. The number of scholarly works in each area over the last decade is shown in Figure \ref{figure:core_issues}.

Language modeling can be viewed in two ways. First, it 
determines which words follow which. By extension, however, this can be viewed as determining what words mean, as individual words are only weakly meaningful, deriving their full value only from their interactions with other words. Morphology is the study of how words themselves are formed. It considers the roots of words and 
the use of prefixes and suffixes, compounds, and other intraword devices,  to display tense, gender, plurality, and a  other linguistic constructs. Parsing  considers which words modify  others, 
forming  constituents, leading to a sentential structure. The area of semantics is the study of what words mean. 
It takes into account the meanings of the individual words and how they relate to and modify others, as well as the context these words appear in and some degree of world knowledge, i.e., ``common sense". 

There is a significant amount of overlap between each of these areas. Therefore, many models analyzed can be classified as belonging in multiple sections. As such, they are discussed in the most relevant sections with logical connections to those other places where they also interact.

\subsection{Language Modeling and Word Embeddings}

\label{section:language_modeling}

Arguably, the most important task in NLP is that of language modeling. Language modeling (LM) is an essential piece of almost any application of NLP. 
Language modeling is the process of creating a model to predict words or simple linguistic components given previous words or components
\cite{Jurafsky2008}. This is 
useful for  applications in which a user types input, 
to provide predictive ability for fast text entry. However, its power and versatility emanate from the fact that it can implicitly capture syntactic and semantic relationships among words or components in a linear neighborhood, making it useful for 
tasks such as machine translation or text summarization. 
Using prediction, such programs are able to generate more relevant, human-sounding sentences. 

\subsubsection{Neural Language Modeling}

\label{section:neural_language_modeling}

A problem with statistical language models was the inability to deal well with synonyms or out-of-vocabulary (OOV) words 
that were not present in the training corpus.
Progress was made in solving the problems  with the introduction of the neural language model \cite{Bengio2003}. While much of NLP took another decade to begin to use ANNs heavily, the LM community immediately took advantage of them, and continued to develop sophisticated models, many of which were summarized by DeMulder et al. \cite{DeMulder2015}.
\subsubsection{Evaluation of Language Models}

\label{section:evaluation_of_langauge_models}

While neural networks have  made breakthroughs in the LM field, it is hard to quantify improvements. 
It is desirable to evaluate language models independently of the applications in which they appear. A number of metrics have been proposed, but no perfect solution has yet been found.
\cite{Iyer1997,Chen1998,Clarkson2001}
The most commonly used metric is perplexity, which is the inverse probability of a test set normalized by the number of words.
Perplexity is a reasonable measurement for LMs trained on the same datasets, but when they are trained on different vocabularies, the metric becomes less meaningful. Luckily, there are several benchmark datasets that are used  in the field, allowing for  comparison. Two such datasets are the Penn Treebank (PTB) \cite{Marcus1993}, and 
the Billion Word Benchmark \cite{Chelba2013}.
\subsubsection{Memory Networks and Attention Mechanisms in Language Modeling}

\label{section:memory_networks_and_attention_in_language_modeling}

Daniluk et al. \cite{Daniluk2017} tested several  networks using  variations of attention mechanisms. The first network  had a simple attention mechanism, which was not fully connected, having a window length of five. 
They hypothesized that using a single value to predict the next token, to encode information for the attentional unit, and to decode the information in the attentional unit hinders a network, as it is difficult to train a single parameter to perform  three distinct tasks simultaneously. Therefore, in the second network, they designed each node to have two outputs: one  to encode and decode the information in the attentional unit, and another to predict the next tokens explicitly. In the third network, they further separated the outputs, using separate values to encode the information entering the attentional unit and decode the information being retrieved from it. 
Tests on a  Wikipedia corpus showed that the attention mechanism improved perplexity compared to the baseline, and that successively adding the second and third parameters led to further increases. 
It was also noted that only the previous five or so tokens carried much value (hence the selection of the window size of five). Therefore, they tested a fourth network which simply used residual connections from each of the previous five units. It was found that this network also provided results comparable to many larger RNNs and LSTMs, suggesting that reasonable results can be achieved using simpler networks. 

Another recent study was done on the usage of residual memory networks (RMNs) for LM \cite{Benes2017}. 
The authors found that residual connections skipping two layers were most effective, followed closely by those skipping a single layer. 
In particular,
a residual connection was present between the first layer and the fourth, as was between the fifth layer and the eighth, and between the ninth and the twelfth. It was found that increasing network depth improved results, but that when using large batch sizes, memory constraints were encountered. Network width was not found to be of particular importance for performance, however, wide networks were found to be harder to train. 
It was found that RMNs are capable of outperforming LSTMs of similar size.
\subsubsection{Convolutional Neural Networks in Language Modeling}

\label{section:CNNs_in_langauge_modeling}

A CNN  used recently in LM replaced the pooling layers  with fully-connected layers \cite{Pham2016}. These layers allowed the feature maps to be reduced to lower dimensional spaces just like the pooling layers. However, whereas any references to location of such features are lost in pooling layers, fully-connected layers somewhat retain this information. 
Three different architectures were implemented: a multilayer perceptron CNN (MLPConv) in which the filters were not simply linear, but instead small MLPs \cite{Lin2013}; a multilayer CNN (ML-CNN) in which multiple convolutional layers were stacked on top of each other; and a combination of these networks called COM, in which kernel sizes for filters varied (in this case they were three and five). 
The results  showed that stacking convolutional layers was  detrimental in LM, but that both MLPConv and COM  reduced  perplexity. 
Combining MLPConv with the varying kernel sizes of COM provided even better results. 
Analysis 
showed that the networks learned specific patterns of words, such as, ``as . . . as". Lastly, this study showed that CNNs can be used to capture long term dependencies in sentences. Closer words were found to be of greatest importance, but words located farther away were  of some significance as well. 
\subsubsection{Character Aware Neural Language Models}

\label{section:character_aware_networks_in_language_modeling}

While most CNNs used in NLP  receive  word embeddings (Section \ref{section:development_of_word_embeddings}) as input, recent networks have analyzed character level input instead. For example, Kim et al. \cite{Kim2016}, unlike previous networks  \cite{Botha2014}, accepted only character level input, rather than combining it with word embeddings. 
A CNN was used to process the character level input  to provide representations of the words. In a similar manner as word embeddings usually are, these representations were then fed into an encoder--decoder pair composed of a highway network (a gated network resembling an LSTM) \cite{Srivastava2015} and an LSTM. They trained the network on the English Penn Treebank, as well as on datasets for Czech, German, Spanish, French, Russian, and Arabic. 
For every non-English language except Russian, the network outperformed previously published results \cite{Botha2014} in both the large and small datasets. 
On the Penn Treebank, results were produced on par with the existing state of the art \cite{Zaremba2014}.
However, the network had only 19 million trainable parameters, which is considerably lower than others. 
Since the network focused on morphological similarities produced by character level analysis, it was more capable than previous models of handling rare words. 
Analysis showed that without the use of highway layers, many words had nearest neighbors that were orthographically similar, but not necessarily semantically similar. 
Additionally, the network was capable of recognizing misspelled words or words not spelled in the standard way (e.g. {\em looooook} instead of {\em look}) and of recognizing out of vocabulary  words. The analysis also showed that the network was capable of identifying prefixes, roots, and suffixes, as well as understanding hyphenated words, making it a robust model.

Jozefowicz et al. \cite{Jozefowicz2016} tested a number of  architectures  
producing character level outputs \cite{Ji2015,Chelba2013,Shazeer2015,Williams2015}. Whereas many of these models had only been tested on small scale language modeling, this study tested them on a large scale, testing them with the Billion Word Benchmark. The most effective model, achieving a state-of-the-art (for single models) perplexity of 30.0 with 1.04 billion trainable parameters (compared to a previous best by a single model of 51.3 with 20 billion parameters \cite{Chelba2013}), was a large LSTM using a character level CNN as an input network. The best performance, however, was achieved using an ensemble 
of ten LSTMs. This ensemble, with a perplexity of 23.7, far surpassed the previous state-of-the-art ensemble \cite{Shazeer2015}, which had a perplexity of 41.0.
\subsubsection{Development of Word Embeddings}

\label{section:development_of_word_embeddings}

Not only do neural language models allow for the prediction of unseen synonymous words, they also allow for modeling the relationships between words \cite{Mikolov2013a, Mikolov2013b}.
Vectors with numeric components, representing individual words, obtained by LM techniques are called embeddings. This  is usually done either by use of Principle Component Analysis  or by capturing internal states in a neural language model. 
(Note that these are not standard LMs, but rather are LMs constructed specifically for this purpose.) 
Typically, word embeddings have between 50 and 300 dimensions. An overused example is that of the distributed representations of the words \textit{king}, \textit{queen}, \textit{man}, and \textit{woman}. If one takes the embedding vectors for each of these words, computation can be performed to obtain highly sensible results. If the vectors representing these words are respectively represented as $\vec{k}$, $\vec{q}$, $\vec{m}$, and $\vec{w}$, it can be observed that $ \vec{k} - \vec{q} \approx \vec{m} - \vec{w} $, which is extremely intuitive to human reasoning. In recent years, word embeddings have been the standard form of input to NLP systems.
\subsubsection{Recent Advances and  Challenges}

\label{section:recent_advances_in_language_modeling}

Language modeling has been evolving on a weekly basis, beginning with the  works of Radford et al. \cite{Radford2018} and Peters et al. \cite{Peters2018}. Radford et al. introduced Generative Pre-Training (GPT) which pretrained a language model based on the Transformer model \cite{Vaswani2017} (Section \ref{section:machine_translation}), learning dependencies of words in sentences and longer segments of text, rather than just the immediately surrounding words. Peters et al. incorporated bi-directionalism  to capture backwards context in addition to the forward context, in their Embeddings from Language Models (ELMo). Additionally, they captured the vectorizations at multiple levels, rather than just the final layer. This allowed for multiple encodings of the same information to be captured, which was empirically shown to boost the performance significantly. 

Devlin et al. \cite{Devlin2018},  added an additional unsupervised training tasks of random masked neighbor word prediction, and next-sentence-prediction (NSP), in which, given a sentence (or other continuous segment of text), another sentence was predicted to either be the next sentence or not. 
These Bidirectional Encoder Representations from Transformers (BERT) were further built upon by Liu et al. \cite{Liu2019b} to create Multi-Task Deep Neural Network (MT-DNN) representations, which are the current state of the art in LM. The model used a stochastic answer network (SAN) \cite{Liu2017, Liu2018b} ontop of a BERT-like model. After pretraining, the model was trained on a number of different tasks before being fine-tuned to the task at hand. 
Using MT-DNN as the LM, they achieved state-of-the-art results on ten out of eleven of the attempted tasks.

While these pretrained models have made excellent headway in ``understanding" language, as is required for some tasks such as entailment inference, it has been hypothesized by some that these models are learning templates or syntactic patterns present within the datasets, unrelated to logic or inference. When new datasets are created removing such patterns carefully, the models do not perform well \cite{McCoy2019}. Additionally, while there has been  recent work on cross-language modeling and universal language modeling, the amount and level of work needs to pick up to address low-resource languages.
\subsection{Morphology}

\label{section:morphology}

Morphology is concerned with finding segments within single words, including roots and stems, prefixes, suffixes, and---in some languages---infixes. Affixes (prefixes, suffixes, or infixes) are used to overtly modify stems for gender, number, person, et cetera. 

Luong et al. \cite{Luong2013} constructed a morphologically aware LM. An RvNN was used to model the morphological structure. A neural language model was then placed on top of the RvNN. The model was trained on the WordSim-353 dataset \cite{Finkelstein2001} and segmentation was performed using Morfessor \cite{Creutz2007}. 
Two models were constructed---one using context and one not. It was found that the model that was insensitive to context over-accounted for certain morphological structures. In particular, words with the same stem were clustered together, even if they were antonyms. The context-sensitive model performed better, noting the relationships between the stems, but also accounting for other features such as the prefix ``un". The model was also tested on several other popular datasets \cite{Miller1991, Rubenstein1965, Huang2012}, significantly outperforming previous embedding models on all.

A good morphological analyzer is often important 
for many NLP tasks. 
As such, one recent study by Belinkov et al. \cite{Belinkov2017} examined the extent to which morphology was learned and used by a variety of neural machine translation models. 
A number of translation models were constructed, all translating from English to French, German, Czech, Arabic, or Hebrew. Encoders and decoders were LSTM-based models (some with attention mechanisms) or character aware CNNs, and the models were trained on the WIT$^3$ corpus \cite{Cettolo2012, Cettolo2016}. The decoders were then replaced with part-of-speech (POS) taggers and morphological taggers, fixing the weights of the encoders to preserve the internal representations. The effects of the encoders were examined as were the effects of the decoders attached during training. The study concluded that the use of attention mechanisms decreases the performance of encoders, but increases the performance of decoders. Furthermore, it was found that character-aware models are superior to others for learning morphology and that the output language affects the performance of the encoders. Specifically, the more morphologically rich the output language, the worse the representations created by the encoders.


Morita et al. \cite{Morita2015} analyzed a new morphological language model for unsegmented languages such as Japanese. They constructed an RNN-based model with a beam search decoder and trained it on  an automatically labeled \cite{Kawahara2006} corpus and on a  manually labeled corpus. The model performed a number of tasks jointly, including morphological analysis, POS tagging, and lemmatization. The model was then tested on the Kyoto Text Corpus \cite{Kawahara2002} and the Kyoto University Web Document Leads Corpus \cite{Hangyo2012}, outperforming all baselines on all tasks.

A recent line of work in  morphology  is  universal morphology. This task considers the relationships between the morphologies of different languages and how they relate to each other, aiming towards the ultimate goal of a single morphological analyzer. However, to the authors' knowledge, there has been only a single study applying deep learning to this area \cite{Dehouck2018}, and even then, only as a supporting task to universal parsing (Section \ref{section:universal_parsing}). For those wishing to apply deep learning to this task, several datasets are already available, including one from a CoNLL shared task \cite{More2018}.

In addition to universal morphology, the development of morphological embeddings, that take into account the structures of words, could aid in multi-language processing. They could possibly be used across cognate languages, which would be valuable when some languages are more resourced than others. In addition, morphological structures may be important in handling specialized languages such as those used in the biomedical literature. Since deep learning has become quite entrenched in NLP, better handling of morphological components is likely to improve performance of overall models. 
\subsection{Parsing}

\label{section:parsing}

Parsing examines how different words and phrases relate to each other within a sentence. There are at least two distinct forms of parsing: constituency parsing and dependency parsing \cite{Jurafsky2008}. In constituency parsing, phrasal constituents are  extracted from a sentence in a hierarchical fashion. 
Dependency parsing  looks  at the relationships between pairs of individual words.

Most recent uses of deep learning in parsing have been in  dependency parsing, within which there exists another major divide in types of solutions. Graph-based parsing constructs a number of parse trees that are then searched to find the correct one. Most graph-based approaches are generative models, in which a formal grammar, based on the natural language, is used to construct the trees \cite{Jurafsky2008}.
More popular in recent years than graph-based approaches have been transition-based approaches that usually construct only one parse tree. While a number of modifications have been proposed, the standard method of transition-based dependency parsing is to create a buffer containing all of the words in the sentence and stack containing only the \textit{ROOT} label. Words are then pushed onto the stack, where connections, known as arcs, are made between the top two items. 
Once dependencies have been determined, words are popped off the stack. The process continues until the buffer is empty and only the \textit{ROOT} label remains on the stack. Three major approaches are used to regulate the conditions in which each of the previously described actions takes place. In the \textit{arc-standard} approach \cite{Nivre2003,Nivre2004}, all dependents are connected to a word before the word is connected to its parent. In the \textit{arc-eager} approach \cite{Nivre2003,Nivre2004}, words are connected to their parents as soon as possible, regardless of whether or not their children are all connected to them. Finally, in the \textit{swap-lazy} approach \cite{Nivre2009}, the arc-standard approach is modified to allow swapping of positions on the stack. This makes the graphing of non-projective edges possible.

\subsubsection{Early Neural Parsing}

\label{section:early_neural_parsing}

One early application of deep learning to NLP, that of Socher et al. \cite{Socher2013a,Socher2013b}, included the use of RNNs with probabilistic context-free grammars (PCFGs) \cite{Fujisaki1991,Jelinek1992}. 
As far as the authors are aware, the first neural model to achieve state-of-the-art performance in parsing was that of Le and Zuidema \cite{Le2014a}. Such performance was achieved on the Penn Treebank for both labeled attachment score (LAS) and unlabeled attachment score (UAS) by using an Inside-Out Recursive Neural Network, which used two vector representations (an inner and an outer) to allow both top-down and bottom-up flows of data. Vinyals et al. \cite{Vinyals2015a} created an LSTM with an attention mechanism in a syntactic constituency parser, which they tested on data from domains different from those of the test data (the English Web Treebank \cite{Petrov2012} and the Question Treebank \cite{Judge2006} as opposed to the Wall Street Journal portion of the Penn Treebank \cite{Marcus1993}), showing that neural models can generalize between domains. Embeddings were first used in dependency parsing by Stenetorp \cite{Stenetorp2013}. 
This approach used an RNN to create a directed acyclic graph. While this model did produce results within 2\% of the state of the art (on the Wall Street Journal portion of the CoNLL 2008 Shared Task dataset \cite{Surdeanu2008}), by the time it reached the end of a sentence, it seemed to have difficulty remembering phrases from early in the sentence.
\subsubsection{Transition-Based Dependency Parsing}

\label{section:transition_based_dependency_parsing}

Chen and Manning \cite{Chen2014} pushed the state of the art in both UAS and LAS on both English and Chinese datasets
on the English Penn Treebank. They accomplished this by using a simple feedforward neural network as the decision maker in a transition-based parser. By doing so they were able to subvert the problem of sparsity persistent in the statistical models. 

Chen and Manning used a simple greedy search, which was replaced by Zhou et al. \cite{Zhou2015} with a beam search, achieving a significant improvement. 
Weiss et al. \cite{Weiss2015} improved upon Chen and Manning's work by using a deeper neural network with residual connections and a perceptron layer placed after the softmax layer. They were able to train on significantly more examples than typical by using tri-training \cite{Li2014}, a process in which potential data samples are fed to two other parsers, and those samples upon which both of the parsers agree are used for training the primary parser. 

Another model was produced using an LSTM instead  of a feedforward network \cite{Dyer2015}. Unlike previous models, this model was given knowledge of the entire buffer and the entire stack and had knowledge of the entire history of transition decisions. This allowed for better predictions, generating state-of-the-art 
on the Stanford Dependency Treebank \cite{deMarneffe2008}, as well as state-of-the-art results on the CTB5 Chinese dataset \cite{Xue2005}. Lastly, Andor et al. \cite{Andor2016} used a feedforward network with global normalization on a number of tasks including part-of-speech tagging, sentence compression, and dependency parsing. State-of-the-art results were obtained on all tasks
on the Wall Street Journal dataset. Notably, their model required significantly less computation than comparable models. 

Much like Stenentorp \cite{Stenetorp2013}, Wang et al. \cite{Wang2018a} used an alternative algorithm to produce directed acyclic graphs, 
for a task called semantic parsing, where deeper relationships between the words are found. 
The task seeks to identify what types of actions are taking place and how words modify each other. In addition to the typical stack and buffer used in transition-based parsing, the algorithm employed a deque. This allowed for the representation of multi-parented words, which although rare in English, are common in many natural languages. Furthermore, it allowed for multiple children of the {\em ROOT} label. In addition to producing said graphs, this work is novel in its use of two new LSTM-based techniques: Bi-LSTM Subtraction and Incremental Tree-LSTM. Bi-LSTM Subtraction built on previous work \cite{Wang2017, Cross2016} to represent the buffer as a subtraction of the vectors from the head and tail of the LSTM, in addition to using an additional LSTM to represent the deque. Incremental Tree-LSTM is an extension of Tree-LSTM \cite{Tai2015}, modified for directed acyclic graphs, by connecting children to parents incrementally, rather than connecting all children to a parent simultaneously. 
The model 
achieved the best published scores at the time for fourteen of the sixteen evaluation metrics used on SemEval-2015 Task 18 (English) \cite{Oepen2015} and SemEval-2016 Task 9 (Chinese) \cite{Che2012}. While deep learning had been applied to semantic parsing in particular domains, such as Question Answering \cite{Yih2014, Krishnamurthy2017}, to the authors' knowledge, this was the first time it was applied in large scale to semantic parsing as a whole.
\subsubsection{Generative Dependency and Constituent Parsing}

\label{section:generative_and_dependency_parsing}

Dyer et al. \cite{Dyer2016} proposed a model that used recurrent neural network grammars for parsing and language modeling. Whereas most approaches take a bottom-up approach to parsing, this took a top-down approach, taking as input the full sentence in addition to the current parse tree. This allowed the sentence to be viewed as a whole, rather than simply allowing local phrases within it to be considered.
This model achieved the then best results in English generative parsing as well as in single sentence language modeling. It also attained results close to the best in Chinese generative parsing.

 Choe and Charniak \cite{Do2016} treated parsing  as a language modeling problem, and used an LSTM  to assign probabilities to the parse trees, achieving state-of-the art. 
Fried et al. \cite{Fried2017} 
wanted to
determine whether the power of the models came from the reranking process or simply from the combined power of two models. They found that while using one parser for producing candidate trees and another for ranking them was superior to a single parser approach, combining two parsers explicitly was preferable. They used two parsers to both select the candidates and rerank them, achieving state-of-the-art results. They extended this model to use three parsers, achieving even better results.
Finally, an ensemble of eight such models (using two parsers) was constructed and achieved the best results 
on Penn Treebank at the time. 

A model created by Dozat and Manning \cite{Dozat2018} used a graph-based approach with a self-attentive network. 
Similarly, Tan et al. \cite{Tan2018} used a self-attentional model for semantic role labeling, a subtask of semantic parsing, achieving excellent results. They experimented with 
recurrent and convolutional replacements to the feed-forward portions of the self-attention mechanism, finding that the feed forward variant had the best performance. Another novel approach  is that of Duong et al. \cite{Duong2018}, who 
used active learning. While not perfect, this is a possible solution to one of the biggest problems in semantic parsing---the availability of data.
\subsubsection{Universal Parsing}

\label{section:universal_parsing}

Much like universal morphology, universal dependency parsing, or universal parsing, is the relatively new task of parsing language using a standardized set of tags and relationships across all languages. While current parsing varies drastically from language to language, this attempts to make it uniform between them, in order to allow for easier processing between and among them. 
Nivre \cite{Nivre2015} discussed the recent development of universal grammar and presented the challenges that lie ahead, mainly the development of tree banks in more languages and the consistency of labeling between tree banks in different (and even the same) languages.
This task has gained traction in large part due to the fact that it has been a CoNLL shared task for the past two years. \cite{Zeman2018} A number of approaches from the 2018 task included using deep transition parsing \cite{Hershcovich2018}, graph-based neural parsing \cite{Ji2018}, and a competitive model which used only a single neural model, rather than an ensemble \cite{Qi2019}. 
The task has begun to be examined outside of CoNLL, with Liu et al. \cite{Liu2018c} applying universal dependencies to the parsing of tweets, using an ensemble of bidirectional LSTM. 

\subsubsection{Remaining Challenges}

Outside of universal parsing, a parsing challenge that needs to be further investigated is the building of syntactic structures without the use of treebanks for training. Attempts have been made using attention scores and Tree-LSTMs, as well as “outside-inside” auto-encoders. If such approaches are successful, they have potential use in many environments, including in the context of low-resource languages and out-of-domain scenarios. While a number of other challenges remain, these are the largest and are expected to receive the most focus. 
\subsection{Semantics}

\label{section:semantics}

Semantic processing involves understanding the meaning of words, phrases, sentences, or documents at some level. 
Word embeddings, such as Word2Vec \cite{Mikolov2013a,Mikolov2013b} and GloVe \cite{Pennington2014}, claim to capture meanings of words, following the Distributional Hypothesis of Meaning \cite{Harris1954}. As a corollary, when vectors corresponding to phrases, sentences, or other components of text are processed using a neural network, a representation that can be loosely thought to be semantically representative is computed compositionally. 
In this section, neural semantic processing research is separated into two distinct areas: Work  on comparing the semantic similarity of two portions of text, and work  on capturing and transferring meaning in high level constituents, particularly sentences.

\subsubsection{Semantic Comparison}

\label{section:semantic_comparison}

One way to test the efficacy of an approach to computing semantics is to see if two similar phrases, sentences or documents, judged by humans to have similar meaning also are judged similarly by a program. 

Hu et al. \cite{Hu2014a} proposed two CNNs to perform a semantic comparison task. The first model, ARC-I, inspired by Bordes et al. \cite{Bordes2014}, used a Siamese network, in which two CNNs sharing weights evaluated two sentences in parallel. In the second network, connections were placed between the two, allowing for sharing before the final states of the CNNs. 
The approach outperformed a number of existing models in tasks in English and Chinese. 

Building on prior work \cite{Socher2011a, Kalchbrenner2014, Hu2014a}, Yin and Sch\"utze \cite{Yin2015} proposed a Bi-CNN-MI (MI for multigranular interaction features), consisting of a pretrained CNN sentence model, a CNN interaction model, and a logistic regressor. 
They modifiied a Siamese network using  Dynamic CNNs  \cite{Kalchbrenner2014} (Section \ref{section:sentence_modeling}). Additionally, the feature maps from each level were used in the comparison, rather than simply the top-level feature maps. They achieved state-of-the-art results on the Microsoft Research Paraphrase Corpus (MSRP) 
\cite{Dolan2004}.

He et al. \cite{He2015} 
constructed feature maps, which were then compared using a ``similarity measurement layer" followed by a fully-connected layer and then a log-softmax output layer  within a CNN. The windows used in the convolutional layers ranged in length from one to four. The network was trained and evaluated on three  datasets: MSRP, the Sentences Involving Compositional Knowledge (SICK) dataset \cite{Marelli2014}, and the Microsoft Video Paraphrase Corpus (MSRVID) \cite{Agirre2012}. State-of-the-art results were achieved on the first and the third. 

Tai et al. concocted a model  using an RvNN with LSTM-like nodes \cite{Tai2015} called 
a Tree-LSTM. Two  variations were examined (constituency-  and dependency-based) and tested on both the SICK dataset and  Stanford Sentiment Treebank \cite{Socher2013a}. The constituency-based model achieved state-of-the-art results on the Stanford Sentiment Treebank and the dependency-based one achieved state-of-the-art results on SICK.

He et al. presented another model \cite{He2016a}, which outperformed that of Tai et al. on SICK. The model formed a matrix of the two sentences  before applying a ``similarity focus layer" and then 
a nineteen-layer CNN followed by dense layers with a softmax output. The similarity focus layer matched semantically similar pairs of words from the input sentences and applied weights to the matrix locations representing the relations between the words in each pair. 
They  also obtained state-of-the-art resuults on MSRVID, SemEval 2014 Task 10 \cite{Agirre2014}, WikiQA \cite{Yang2015}, and TreeQA \cite{Wang2007} datasets. 
\subsubsection{Sentence Modeling}

\label{section:sentence_modeling}

Extending from neural language modeling, 
sentence modeling attempts to capture the meaning of sentences in vectors. Taking this a step further are models, such as that of Le and Mikolov \cite{Le2014b}, which attempt to model paragraphs or larger bodies of text in this way. 

Kalchbrenner et al. \cite{Kalchbrenner2014} generated representations of sentences using a dynamic convolutional neural network (DCNN), which used a number of  filters and dynamic $k$-max pooling layers. 
Due to dynamic pooling, features of different types and lengths could be identified in sentences with varying structures without padding of the input. This allowed not only short-range dependencies, but also long-range dependencies to be identified. The DCNN was tested in  applied tasks that require  semantic understanding. It outperformed all comparison models in predicting sentiment of movie reviews in the Stanford Sentiment Treebank \cite{Socher2013b} and in identification of sentiment in tweets \cite{Go2009}. It was also one of the top performers in classifying types of questions using the TREC database \cite{Li2002}.

Between their requirement for such understanding and their ease of examination due to the typical encoder--decoder structure they use, 
neural machine translation (NMT) systems (Section \ref{section:machine_translation}) are splendid testbeds for researching internal semantic representations. 
Poliak et al. \cite{Poliak2018a} trained encoders  on four different language pairs: English and Arabic, English and Spanish, English and Chinese, and English and German. 
The decoding classifiers were trained on four distinct datasets: Multi-NLI \cite{Williams2017}, which is an expanded version of SNLI \cite{Nangia2017}, as well as three recast datasets from the JHU Decompositional Semantics Initiative \cite{White2017} (FrameNet Plus or FN+ \cite{Pavlick2015},  Definite Pronoun Resolution or DPR \cite{Rahman2012}, and Semantic Proto-Roles or SPR \cite{Reisinger2015}). None of the results were particularly strong, although they were strongest in SPR. This led  to the conclusion that NMT models do a poor job of capturing paraphrased information and fail to capture inferences that help in anaphora resolution. 
(e.g. resolving gender). 
They did, however, find that the models  learn  about proto-roles (e.g. who or what is the recipient of an action). A concurrent work \cite{Poliak2018b} analyzed the quality of many datasets used for natural language inference.

Herzig and Berant \cite{Herzig2017} found that training semantic parsers on a single domain, as is often done, is less effective than training across many domains. This conclusion was drawn after 
testing three LSTM-based models. The first model was a one-to-one model, in which a single encoder and single decoder were used, requiring the network itself to determine the domain of the input. In the second model, a many-to-many model, a decoder was used for each domain, as were two encoders: the domain specific encoder and a multidomain encoder. The third model was a one-to-many model, using a single encoder, but separate decoders for each domain. Each model was trained on the ``OVERNIGHT" dataset \cite{Wang2015}. Exceptional results were achieved for all models, with a state-of-the-art performance exhibited by the one-to-one model.

Similar conclusions were drawn 
by Brunner et al. \cite{Brunner2018}. 
who created several LSTM-based encoder--decoder networks, and analyzed the embedding vectors produced. A single encoder accepting English sentences as input was used, as were four different decoders. The first such decoder was a replicating decoder, which 
reproduced the original English input. The second and third decoders 
translated the text into German and French. Finally, the fourth decoder was a POS tagger. Different combinations of decoders were used; one model had only the replicating decoder while others had two, three, or all four. Sentences of fourteen different structures from the EuroParl dataset \cite{Koehn2005} were used to train the networks. A set of test sentences were then fed to the encoders and their output analyzed. In all cases, fourteen clusters were formed, each corresponding to one of the sentence structures. 
Analysis
showed that adding more decoders led to more correct and more definitive clusters. In particular, using all four of the decoders led to  zero error. Furthermore, the researchers confirmed a hypothesis 
that just as logical arithmetic can be performed on word embeddings, so can it be performed on sentence embeddings. 

\subsubsection{Semantic Challenges}

In addition to the challenges already mentioned,  researchers believe that being able to solve tasks well does not indicate actual understanding. Integrating deep networks with general word-graphs (e.g. WordNet \cite{Miller1995}) or knowledge-graphs (e.g. DBPedia \cite{Auer2007}) may be able to endow a sense of understanding. Graph-embedding is an active area of research \cite{Wang2017b}, and work on integrating language-based models and graph models has only recently begun to take off, giving hope for better machine understanding.
\subsection{Summary of Core Issues}

\label{section:summary_of_core_issues}

Deep learning has generally performed very well, surpassing existing states of the art in many individual core NLP tasks, and has thus created the foundation on which useful natural language applications can and are being built. However, it is clear from examining the research reviewed here that natural language is an enigmatically complex topic, with myriad core or basic tasks, of which deep learning has only grazed the surface. It is also not clear how architectures for ably executing individual core tasks can be synthesized to build a common edifice, possibly a much more complex distributed neural architecture, to show competence in multiple or ``all" core tasks.  More fundamentally, it is also not clear, how mastering of basic tasks, may lead to superior performance in applied tasks, which are the ultimate engineering goals, especially in the context of building effective and efficient deep learning models. 
Many, if not most, successful deep learning architectures for applied tasks, discussed in the next section, seem to forgo explicit architectural components for core tasks, and learn such tasks implicitly.
Thus, some researchers argue that the relevance of the large amount of work on core issues is not fully justified, while others argue that further extensive research in such areas is necessary to better understand and develop systems which more perfectly perform these tasks, whether explicitly or implicitly.


\section{Applications of Natural Language Processing Using Deep Learning}

\label{section:applications}

\begin{figure*}

    \centering
    
    \includegraphics[width=7.00in]{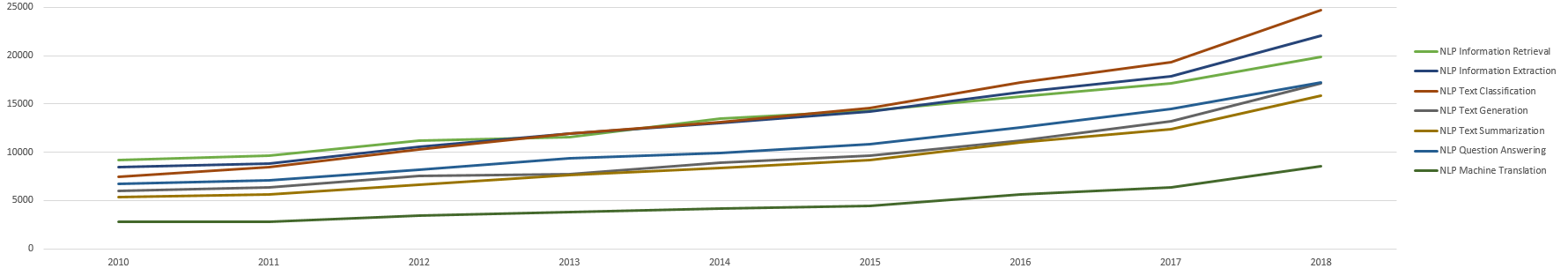}
    
    \caption{Publication Volume for Applied Areas of NLP. All areas of applied natural language processing discussed have witnessed growth in recent years, with the largest growth occurring in the last two to three years.}
    
    \label{figure:applications}

\end{figure*}

While the study of core areas of NLP is important to understanding how neural models work, it is meaningless in and of itself from an engineering perspective, which values applications that benefit humanity, not pure philosophical and scientific inquiry. 
Current approaches to solving several immediately useful NLP tasks  are summarized here. Note that the issues included here are only those involving the processing of text, not the processing of verbal speech. Because speech processing \cite{Hinton2012,Graves2013} requires expertise on several other topics including acoustic processing, it is generally considered another field of its own, sharing many commonalities with the field of NLP. 
The number of studies in each discussed area over the last decade is shown in Figure \ref{figure:applications}

\subsection{Information Retrieval}

\label{section:information_retrieval}

The purpose of Information Retrieval (IR) systems is to help people find the right (most useful) information in the right (most convenient) format at the right time (when they need it) \cite{Kenter2017}. Among many  issues in IR, a primary problem that needs  addressing pertains to ranking documents with respect to a query string in terms of relevance scores for ad-hoc retrieval tasks, similar to what happens in a search engine. 


 Deep learning models for ad-hoc retrieval match texts of queries to texts of documents to obtain relevance scores. Thus, such models have to focus on producing representations of the interactions among individual words in the query and the documents. Some representation-focused approaches build deep learning models to produce good representations for the texts and then match the representations straightforwardly \cite{Huang2013,Hu2014a,Shen2014}, whereas interaction-focused approaches first build local interactions directly, and then use deep neural networks to learn how the two pieces of text match based on word interactions \cite{Hu2014a,Lu2013,Pang2016}. When matching a long document to a short query, the relevant portion can potentially occur anywhere in the long document and may also be distributed, thus, finding how each word in the query relates to portions of the document is helpful.

Mindful of the specific needs for IR, Guo et al. \cite{Guo2016b} built a neural architecture called DRMM, enhancing an interaction-focused model that feeds quantized histograms of the local interaction intensities to an MLP  for matching. In parallel, the query terms go through a small sub-network on their own to establish term importance and term dependencies. The outputs of the two parallel networks are mixed at the top  so that the relevance of the document to the query can be better learned. DRMM achieved state-of-the-art performance for its time.

 Most current neural IR models are not end-to-end relevance rankers, but are  re-rankers for documents a first-stage efficient traditional ranker has deemed relevant to a query. The representations the neural re-rankers  learn  are dense  for both documents and queries, i.e., most documents in a collection seem to be relevant to a query, making it impossible to use such ANNs for ranking an entire collection of documents. In contrast, Zamani et al. \cite{Zamani2018} presented a standalone neural ranking model called SNRM\_PRF, that learned sparse representations for both queries and documents, mimicking what traditional approaches do. Since queries are much shorter than documents and queries contain much less information than documents, it makes sense for query representations to be denser. 
 This was achieved by using, during training, a sparsity objective combined with hinge loss. In particular, an $n$-gram  representation for queries and documents was used.
It passed the embedding of each word separately through an individual MLP and performed average pooling on top. During training, the approach   used pseudo-relevant documents obtained by retrieving documents using existing models like TF-IDF and BM25, because of the lack of enough correctly labeled documents to train large ANN models. The approach 
created a 20,000 bit long inverted index for each document using the trained network, just like a traditional end-to-end approach. For retrieval, a dot product was computed between query and document representations to obtain the retrieval relevance score.  The SNRM\_PRF system obtained the best metrics (measured by MAP, P@20, nDCG@20, and Recall) across the board for two large datasets, Robust and ClueWeb. 

 MacAveney et al. \cite{MacAvaney2019} extracted query term representations from two pre-trained contextualized language models, ELMo \cite{Peters2018} and BERT \cite{Devlin2018}, and used the representations to augment three existing competitive neural ranking architectures for ad-hoc document ranking, one of them being DRMM \cite{Guo2016b}. They also presented a joint model that combined BERT's classification vector with these  architectures to get benefits from both approaches. MacAveney's system called CEDR (Contextualized Embeddings for Document Ranking) improved performance of all three prior models, and produced state-of-the-art results using BERT's token representations.
\subsection{Information Extraction}

\label{section:information_extraction}

Information extraction 
extracts explicit or implicit information from text. The outputs of systems 
vary, but often the extracted data and the relationships within it are saved in relational databases \cite{Cowie1996}.
Commonly extracted information includes named entities and relations, events and their participants, temporal information, and tuples of facts.

\subsubsection{Named Entity Recognition}

\label{section:named_entity_recogntion}

Named entity recognition (NER) refers to the identification of proper nouns as well as information such as dates, times, prices, and product IDs. 
The multi-task approach of Collobert et al. \cite{Collobert2011} included the task, although no results were reported. In their approach, a simple feedforward network was used, having a context with a fixed sized window around each word. Presumably, this made it difficult to capture long-distance relations between words. 

LSTMs were first used for NER by Hammerton \cite{Hammerton2003}. The model, which was ahead of its time, had a small network due to the lack of available computing power at the time. Additionally, sophisticated numeric vector models for words were not yet available. Results  were slightly better than the baseline for English and much better than the baseline for German. Dos Santos et al. \cite{DosSantos2015} used a  deep neural network  architecture, known as CharWNN, which jointly used word-level and character-level inputs to perform sequential classification. 
In this study, a number of experiments were performed using the HAREM I annotated Portuguese corpus \cite{Santos2006}, and the SPA CoNLL2002 annotated Spanish corpus \cite{Carreras2002}. For the Portuguese corpus, CharWNN outperformed the previous state-of-the-art system 
across ten named entity classes. It also achieved state-of-the-art performance in Spanish. The authors noted that when used alone, neither word embeddings nor character level embeddings worked. 
This revalidated a fact long-known: 
Joint use of word-level and character-level features is important to effective NER performance.

Chiu and Nichols \cite{Chiu2015} 
used a bidirectional LSTM  with a character-level CNN resembling those used by dos Santos et al. \cite{DosSantos2015}. 
Without using any private lexicons, detailed information about linked entities, or 
produce state-of-the-art results on the CoNLL-2003 \cite{Tjong2003} and OntoNotes \cite{Hovy2006,Pradhan2013} datasets.

Lample et al. \cite{Lample2016} developed an architecture based on bidirectional LSTMs and conditional random fields (CRFs). The model used both character-level inputs and word embeddings. The inputs were combined and then fed to a bidirectional LSTM, whose outputs were in turn fed to a layer that performed CRF computations \cite{Lafferty2001}. The model, when trained using dropout, obtained state-of-the-art performance in both German and Spanish. The LSTM-CRF model was also very close in both English and Dutch. The  claim of this study was that state-of-the-art results were achieved without the use of any hand-engineered features or gazetteers.

Akbik et al. \cite{Akbik2018} achieved state-of-the-art performance in German and English NER  using a pre-trained bidirectional character language model. They retrieved for each word a contextual embedding that they passed into a BiLSTM-CRF sequence labeler to perform NER. 

\subsubsection{Event Extraction}

\label{section:event_extraction}

Event extraction is concerned with identifying words or phrases that refer to the occurrence of events, along with participants such as agents, objects,  recipients, and times of occurrence. Event extraction usually deals with four sub-tasks: identifying event mentions, or phrases that describe events; identifying event triggers, which are the main words---usually verbs or gerunds
---that specify the occurrence of the events; identifying arguments of the events; and identifying arguments' roles in the events.
 
Chen et al. \cite{Chen2015} 
argued that CNNs that use max-pooling are likely to capture only the most important information in a sentence, and as a result, might miss valuable facts when considering sentences that refer to several events. To address this drawback, they divided the feature map into three parts, and instead of using one maximum value, kept the maximum value of each part. 
In the first stage, they classified each word  as either being a trigger word or non-trigger word. If triggers were found, the second stage aligned the roles of arguments. 
Results showed that this approach significantly outperformed other state-of-the-art methods of the time. The following year, Nguyen et al. \cite{Nguyen2016} used an RNN-based encoder--decoder pair to identify event triggers and roles, exceeding earlier results.
Liu et al. \cite{Liu2019c} presented a latent variable neural model to induce event schemas and extract open domain events, achieving best results on a dataset they created and released.

\subsubsection{Relationship Extraction}

\label{section:relationship_extraction}

Another important type of information extracted from text is that of relationships. These may be possessive, antonymous or synonymous relationships, or more natural,  familial or geographic, relationships. 
The first deep learning approach was that of Zeng et al. \cite{Zeng2014}, who used a simple CNN to classify a number of relationships between elements in sentences. Using only two layers, a window size of three, and word embeddings with only fifty dimensions they attained better results than any prior approach.
Further work, by Zheng et al. \cite{Zheng2017}, used a bidirectional LSTM and a CNN for relationship classification as well as entity recognition. More recently, Sun et al. \cite{Sun2018} used an attention-based GRU model with a copy mechanism. This network was novel in its use of a data structure known as a coverage mechanism \cite{Tu2016}, which helped  ensure that all important information was extracted the correct number of times.
Lin et al. \cite{Lin2019} achieve state-of-the-art performance in clinical temporal relation extraction using the pre-trained BERT \cite{Devlin2018} model with supervised training on a biomedical dataset. 
\subsection{Text Classification}

\label{section:classification}

Another classic application for NLP is text classification, or the assignment of free-text documents to predefined classes. Document classification has numerous applications. 

Kim \cite{Kim2014} was the first to use pretrained word vectors in a CNN for sentence-level classification. Kim's work  was motivating, and showed that simple CNNs, with one convolutional layer followed by a dense layer with dropout and  softmax output, could achieve excellent results on multiple benchmarks using little hyperparameter tuning. The CNN models proposed  were able to improve upon the state of the art on 4 out of 7 different tasks cast as sentence classification, including sentiment analysis and question classification. Conneau et al. \cite{Conneau2017} later showed that networks that employ a large number of convolutional layers work well for document classification. 

Jiang \cite{Jiang2018} used a hybrid architecture combining a deep belief network \cite{Hinton2006} and softmax regression \cite{Sutton1998}. (A deep belief network is a feedforward network where pairs of hidden layers are designed to resemble  restricted Boltzmann machines  \cite{Smolensky1986}, which are trained using unsupervised learning and are designed to increase or decrease dimensionality of data.) This was achieved by making passes over the data using forward and backward propagation many times until a minimum engery-based loss was found. This process was independent of the labeled or classification portion of the task, and was therefore initially trained without the softmax regression output layer. Once both sections of the architecture were pretrained, they were  combined and trained like a regular deep neural net with backpropagation and quasi-Newton methods \cite{Fletcher2013}.

Adhikari et al. \cite{Adhikari2019} used BERT \cite{Devlin2018} to obtain state-of-the-art classification results on four document datasets. 

While deep learning is promising for many areas of NLP, including text classification, it is not necessarily the end-all-be-all, and many hurdles are still present.  Worsham and Kalita \cite{Worsham2018} found that for the task of classifying long full-length books by genre,   gradient boosting trees are superior to neural networks, including both CNNs and LSTMs. 

\subsection{Text Generation}

\label{section:text_generation}

Many NLP tasks require the generation of human-like language. Summarization and machine translation convert one text to another in a sequence-to-sequence (seq2seq) fashion. Other tasks, such as image and video captioning and automatic weather and sports reporting, convert non-textual data to text. Some tasks, however, produce text without any input data to convert (or with only small amounts used as a topic or guide). These tasks include poetry generation, joke generation, and story generation.

\subsubsection{Poetry Generation}

\label{section:poetry_generation}

Poetry generation is arguably the hardest of the  generation subtasks, as in addition to producing creative content, the content must be delivered in an aesthetic manner, usually following a specific structure. As with most tasks requiring textual output, recurrent models are the standard. However, while recurrent networks are great at learning internal language models, they do a poor job of producing structured output or adhering to any single style. Wei et al. \cite{Wei2018} addressed the style issue by training  using particular poets and controlling for style in Chinese poetry. They found that with enough training data, adequate results could be achieved. The structure problem was addressed by Hopkins and Kiela \cite{Hopkins2017}, who generated rhythmic poetry by  training the network on only a single type of poem to ensure produced poems adhered to a single rhythmic structure. Human evaluators judged poems produced  to be of lower quality than, but indistinguishable from, human produced poems. 

Another approach to poetry generation, beginning this year, has been to use pretrained language models. Specifically, Radford et al.'s GPT-2 model \cite{Radford2019}, the successor of the GPT model  (Section \ref{section:recent_advances_in_language_modeling}) has been used. Radford et al. hypothesised that alongside sequence-to-sequence learning and attention, language models can inherently start to learn text generation while training over a vast dataset. As of late 2019, these pre-trained GPT-2 models are arguably the most effective and prolific neural natural language generators. 
Bena and Kalita \cite{Bena2019} used the 774 million parameter GPT-2 model to generate high-quality poems in English, demonstrating and eliciting emotional response in readers. (Two other models are available: 355 million parameters, and as of Novemeber 2019, 1.5 billion parameters.) Tucker and Kalita \cite{Tucker2019} generated poems in several languages---English, Spanish, Ukrainian, Hindi, Bengali, and Assamese---using the 774 M model as well. This study provided astonishing results in the fact that GPT-2 was pre-trained on a large English corpus, yet with further training on only a few hundred poems in another language, it turns into a believable generator in that language, even for poetry.

\subsubsection{Joke and Pun Generation}

\label{section:humor_generation}

Another area which has received little attention is the use of deep learning for joke and pun generation. Yu et al. \cite{Yu2018} generated homographic puns (puns which use multiple meanings of the same written word) using a small LSTM. The network  produced sentences in which ambiguities were introduced by words with multiple meanings, although it did a poor job of making the puns humorous. The generated puns were classified by human evaluators as machine generated a majority of the time. The authors noted that training on pun data alone is not sufficient for generating good puns. 
Ren and Yang \cite{Ren2017} used an LSTM to generate jokes, training on two datasets, one of which was a collection of short jokes from Conan O'Brien. Since many of these jokes pertain to current events, the network was also trained on a set of news articles. This gave context to the example jokes. 
Chippada and Saha \cite{Chippada2018} generated jokes, quotes, and tweets using the same neural network, using an additional input to specify which should be produced. It was found that providing more general knowledge of other types of language, and examples of non-jokes, increased the quality of the jokes produced.

\subsubsection{Story Generation}

\label{section:story_generation}

While poetry and especially humor generation have not gained much traction, story generation has seen a recent rise in interest. Jain et al. \cite{Jain2017} used RNN variants with attention to produce short stories from ``one-liner" story descriptions. 
Another recent study of interest is that by Peng et al. \cite{Peng2018}, who used LSTMs to generate stories, providing an input to specify whether the story should have a happy or sad ending. Their model successfully did so while at the same time providing better coherence than non-controlled stories.
More recent attempts at the task have used special mechanisms focusing on the ``events" (or actions) in the stories \cite{Martin2018} or on the entities (characters and important objects) \cite{Clark2018}. Even with such constraints, generated stories generally become incoherent or lose direction rather shortly. Xu et al. \cite{Xu2018} addressed this by using a ``skeleton" based model to build general sentences and fill in important information. This did a great job of capturing only the most important information, but still provided only modest end results in human evaluation. Drissi et al. \cite{Drissi2018} followed a similar approach. 

The strongest models to date focus on creating high level overviews of stories before breaking them down into smaller components to convert to text. Huang et al. \cite{Huang2018} generated short stories from images using a two-tiered network. The first constructed a conceptual overview while the second converted the overview into words. Fan et al. \cite{Fan2018} used a hierarchical approach, based on CNNs, which beat out the non-hierarchical approach in blind comparison by human evaluators. 
Additionally, they found that self attention leads to better perplexity. They also developed a fusion model with a pretrained language model, leading to greater improvements. These results concur with those of an older study by Li et al. \cite{Li2015} who read documents in a hierarchical fashion and reproduced them in hierarchical fashion, achieving great results.

\subsubsection{Text Generation with GANs}

\label{section:text_generation_with_GANs}

In order to make stories seem more human-like, He et al. \cite{Lin2017} used GANs (generative adversarial networks) to measure human-likeness of generated text, forcing the network toward more natural reading output. Generative Adversarial Networks are based on the concept of a minimax two-player game, in which a generative network and a discriminative network are designed to work against each other with the discriminator attempting to determine if examples are from the generative network or the training set, and the generator trying to maximize the number of mistakes made by the discriminator. RankGAN, the GAN used in the study, measured differences in embedding space, rather than in output tokens. This meant the story content was evaluated more directly, without respect to the specific words and grammars used to tell it. Rather than simply using standard metrics and minimizing loss, Tambwekar et al. \cite{Tambwekar2018} used reinforcement learning to train a text generation model. This taught the model to not only attempt to optimize metrics, but to generate stories that humans evaluated to be meaningful. Zhange et al. \cite{Zhang2017} used another modified GAN, referred to as textGAN, for text generation, employing an LSTM generator and a CNN discriminator, achieving a promising BLEU score and a high tendency to reproduce realistic-looking sentences. Generative adversarial networks have seen increasing use in text generation recently 
\cite{Chen2018b, Guo2018}.

\subsubsection{Text Generation with VAEs}

\label{section:text_generation_with_VAEs}

Another interesting type of network is the variational autoencoder (VAE) \cite{Kingma2013}. While GANs attempt to produce output indistinguishable (at least to the model's discriminator) from actual samples, VAEs attempt to create output similar to samples in the training set \cite{Doersch2016}. Several recent studies have used VAEs for text generation \cite{Serban2017, Hu2018}, including Wang et al. \cite{Wang2019}, who adapted it by adding a module for learning a guiding topic for sequence generation, producing good results.

\subsubsection{Summary of Text Generation}

\label{section:text_generation_summary}

Humor and poetry generation are still understudied topics. As machine generated texts improve, the desire for more character, personality, and color in the texts will almost certainly emerge. Hence, it can be expected that research in these areas will increase.

While story generation is improving, coherence is still a major problem, especially for longer stories. This has been addressed in part, by Haltzman et al., who \cite{Holtzman2019} have proposed ``nucleus sampling" to help counteract this problem, performing their experiments using the GPT-2 model.

In addition to issues with lack of creativity and coherence, creating metrics to measure any sort of creative task is difficult, and therefore, human evaluations are the norm, often utilizing Amazon's Mechanical Turk. However, recent works have proposed metrics that make a large step toward reliable automatic evaluation of generated text \cite{Clark2019, Hashimoto2019}. In addition to the more creative tasks surveyed here, a number of others were previously discussed by Gatt and Krahmer \cite{Gatt2018}. The use of deep learning for image captioning has been surveyed very recently \cite{Hossain2019, Liu2019a}, and tasks that generate text given textual inputs are discussed in the following subsections.

\subsection{Summarization}

\label{section:summarization}

Summarization  finds elements  of interest in documents in order to produce an encapsulation of the most important content. There are two primary types of summarization: extractive and abstractive. The first focuses on sentence extraction, simplification, reordering, and concatenation to relay the important information  in documents using text taken directly from the documents.
Abstractive summaries rely on expressing documents' contents through generation-style abstraction, possibly using words never seen in the documents \cite{Jurafsky2008}.

Rush et al. \cite{Rush2015} introduced deep learning to summarization, using a feedforward neural networrk. The language model used an  encoder and a generative beam search decoder. The initial input was given directly to both the language model and the convolutional attention-based encoder, which determined contextual importance surrounding the summary sentences and phrases. The performance of the model was comparable to other state-of-the-art models of the time.

As in other areas, attention mechanisms have improved performance of encoder--decoder models. Krantz and Kalita \cite{Krantz2018} compared various attention models for abstractive summarization. A state-of-the-art approach developed by Paulus et al. \cite{Paulus2017} used a multiple intra-temporal attention encoder mechanism that considered not only the input text tokens, but also the output tokens used by the decoder for previously generated words. They also used similar hybrid cross-entropy loss functions to those proposed by Ranzato et al. \cite{Ranzato2015}, which led to decreases in training and execution by orders of magnitude. Finally, they recommended using strategies seen in reinforcement learning to modify gradients and reduce exposure bias, which has been noted in models trained exclusively via supervised learning. The use of attention also boosted accuracy in the fully convolutional model proposed by Gehring et al. \cite{Gehring2017}, who implemented an attention mechanism for each layer.

Zhang et al. \cite{Zhang2019} proposed an encoder-decoder framework, which  generated an output sequence based on an input sequence in a two-stage manner. They encoded the input sequence  using BERT \cite{Devlin2018}. The decoder had two stages. In the first stage,  a Transformer-based decoder generated a draft output sequence. In the second stage, they masked each word of the draft sequence and fed it to  BERT, and then by combining the input sequence and the draft representation generated by BERT, they used a Transformer-based decoder to predict the refined word for each masked position. Their model achieved state-of-the-art performance on the CNN/Daily Mail and New York Times datasets.

\subsection{Question Answering}

\label{section:question_answering}

Similar to summarization and information extraction, question answering (QA) gathers relevant words, phrases, or sentences from a document. QA  returns this information in a coherent fashion in response to a request.
Current  methods resemble those of summarization.

Wang et al. \cite{Wang2017} used a gated attention-based recurrent network to match the question with an answer-containing passage. A self-matching attention mechanism was used to refine the machine representation by mapping the entire passage. Pointer networks were used to predict the location and boundary of an answer. These networks used attention-pooling vector representations of passages, as well as the words being analyzed, to model the critical tokens or phrases necessary. 

Multicolumn CNNs  were used by Dong et al. \cite{Dong2015} to automatically analyze questions from multiple viewpoints. Parallel networks were used to extract pertinent information from input questions. Separate networks were used to find context information and relationships and to determine which forms of answers should be returned. The output of these networks was  combined and used to rank possible answers. 

Santoro et al. \cite{Santoro2017}  used relational networks (RNs) for summarization. First proposed by Raposo et al. \cite{Raposo2017}, RNs are  built upon an MLP architecture, with focus on relational reasoning, i.e. defining relationships among entities in the data. These feedforward networks implement a similar function among all pairs of objects in order to aggregate correlations among them. For input, the RNs took final LSTM representations of document sentences. These inputs were further paired with a representation of the information request given \cite{Santoro2017}. 

BERT \cite{Devlin2018} achieved state of theart in QA experiments on SQuAD 1.1 and SQuAD 2.0 datasets. Yang et al. \cite{Yang2019}  demonstrate an end-to-end question answering system that integrates BERT with the open-source Anserini information retrieval toolkit. This system is able to identify answers from a large corpus of Wikipedia articles in an end-to-end fashion, obtaining best results on a standard benchmark test collection. 
\subsection{Machine Translation}

\label{section:machine_translation}

Machine translation (MT) is the quintessential application of NLP. It involves the use of mathematical and algorithmic techniques to translate documents in one language to another. Performing effective translation is intrinsically onerous even for humans, requiring  proficiency  in areas such as morphology, syntax, and semantics, as well as an adept understanding and discernment of cultural sensitivities, for both of the languages (and associated societies) under consideration \cite{Jurafsky2008}.

The first attempt at neural machine translation (NMT) was that by Schwenk \cite{Schwenk2012}, although neural models had previously been used for the similar task of transliteration, converting certain parts of text, such as proper nouns, into different languages \cite{Deselaers2009}. Schwenk used a feed-forward network with seven-word inputs and outputs, padding and trimming when necessary. The ability to translate from a sentence of one length to a sentence of another length came about with the introduction of encoder-decoder models.

The first use of such a model, by Kalchbrenner and Blumson \cite{Kalchbrenner2013}, stemmed from the success of continuous recurrent representations in capturing syntax, semantics, and morphology \cite{Collobert2008} in addition to the ability of RNNs to build robust language models \cite{Mikolov2010}. This original NMT encoder-decoder model used a combination of generative convolutional and recurrent layers to encode and optimize a source language model and cast this into a target language. The model was quickly reworked and further studied by Cho et al. \cite{Cho2014b} and numerous novel and effective advances to this model have since been made \cite{Bahdanau2014, Sutskever2014}. Encoder-decoder models have continuously defined the state of the art, being expanded to contain dozens of layers, with residual connections, attention mechanisms, and even residual attention mechanisms allowing the final decoding layer to attend to the first encoding layer \cite{Wu2016b}. State-of-the-art results have also been achieved by using numerous convolutional layers in both the encoder and decoder, allowing information to be viewed in several hierarchical layers rather than a multitude of recurrent steps \cite{Gehring2017}. Such derived models are continually improving, finding answers to the shortcomings of their predecessors and overcoming any need for hand engineering \cite{Britz2017}. Recent progress includes effective initialization of decoder hidden states, use of conditional gated attentional cells, removal of bias in embedding layers, use of alternative decoding phases, factorization  of embeddings, and test time use of the beam search algorithm \cite{Sennrich2017, Klein2017}.

The standard initialization for the decoder state is that proposed by Bahdanau et al. \cite{Bahdanau2014}, using the last backward encoder state. However, as noted by Britz et al. \cite{Britz2017}, using the average of the embedding or annotation layer seems to lead to the best translations. Gated recurrent cells have been the gold standard for sequence-to-sequence tasks, a variation of which is a conditional GRU (cGRU) \cite{Sennrich2017}, most effectively utilized with an attention mechanism. A cGRU cell consists of three key components: two GRU transition blocks and an attention mechanism between them. These three blocks combine the previous hidden state, along with the attention context window to generate the next hidden state. Altering the decoding process \cite{Bahdanau2014} from  \textit{Look} at input, \textit{Generate} output token, \textit {Update} hidden representation to a process of \textit{Look, Update, Generate} can simplify the final decoding. Adding further source attributes such as morphological segmentation labels, POS tags, and syntactic dependency labels  improves models, and concatenating or factorizing these with embeddings increases robustness further \cite{Sennrich2016, Sennrich2017}. For remembering long-term dependencies, vertically stacked recurrent units have been the standard, with the optimum number of layers having been determined to be roughly between two and sixteen \cite{Britz2017}, depending on the desired input length as well as the presence and density of residual connections. At test time, a beam search algorithm can be used beside the final softmax layer for considering multiple target predictions in a greedy fashion, allowing the best predictions to be found without looking through the entire hypothesis space \cite{Klein2017}.

In a direction diverging from previous work, Vaswani et al. \cite{Vaswani2017, Ahmed2017} proposed discarding the large number of recurrent and convolutional layers and instead focusing exclusively on attention mechanisms to encode a language globally from input to output. Preferring such "self-attention" mechanisms over traditional layers is motivated by the following three principles: reducing the complexity of computations required per layer, minimizing sequential training steps, and lastly, abating the path length from input to output and its handicap on the learning of the long-range dependencies which are necessary in many sequencing tasks \cite{Hochreiter2001}. Apart from increased accuracy across translation tasks, self-attention models allow more parallelization throughout architectures, decreasing the training times and minimizing necessary sequential steps. At time of writing, the state-of-the-art model generating the best results for English to German and English to French on the IWSLT (International Workshop on Spoken Language Translation) 2014 test corpus \cite{Cettolo2014} is that of Medina and Kalita \cite{Medina2018}, which modified the model proposed by Vaswani to use parallel self-attention mechanisms, rather than stacking them as was done in the original model. In addition to improving BLEU (Bilingual Evaluation Understudy) scores \cite{Papineni2002}, this also reduced training times. Ghazvininejad et al. \cite{Ghazvininejad2019} recently applied BERT to the machine translation task, using constant-time models. They were able to achieve relatively competitive performance in a fraction of the time. Lample et al. \cite{Lample2019} attained state-of-the-art results, performing unsupervised machine translation using multiple languages in their language model pretraining.

Several of the recent state-of-the-art models were examined by Chen et al. \cite{Chen2018}. The models were picked apart to determine which features were truly responsible for their strength and to provide a fair comparison. Hybrid models were then created using this knowledge, and incorporating the best parts of each previous model, outperforming the previous models. In addition to creating two models with both a self-attentive component and a recurrent component (in one model they were stacked, in the other parallel), they determined four techniques which they believe should always be employed, as they are crucial to some models, at best, and neutral to all models examined, at worst. These are label smoothing, multi-head attention, layer normalization, and synchronous training. Another study, by Denkowski et al. \cite{Denkowski2017}, examined a number of other techniques, recommending three: using Adam optimization, restarting multiple times, with learning rate annealing; performing subword translation; and using an ensemble of decoders. Furthermore, they tested a number of common techniques on models that were strong to begin, and determined that three of the four provided no additional benefits to, or actually hurt, the model, those three being lexicon-bias (priming the outputs with directly translated words), pre-translation (using translations from another model, usually of lower quality, as additional input), and dropout. They did find, however, that data-bootstrapping (using phrases that are parts of training examples as additional independent smaller samples) was advantageous even to models that are already high-performing. They recommended that future developments be tested on top performing models in order to determine their realm of effectiveness.

In addition to studies presenting recommendations, one study has listed a number of challenges facing the field \cite{Koehn2017}. While neural machine translation models are superior to other forms of statistical machine translation models (as well as rule-based models), they require significantly more data, perform poorly outside of the domain in which they are trained, fail to handle rare words adequately, and do not do well with long sentences (more than about sixty words). Furthermore, attention mechanisms do not perform as well as their statistical counterparts for aligning words, and beam searches used for decoding only work when the search space is small. Surely these six drawbacks will be, or in some cases, will continue to be, the focus of much research in the coming years. Additionally, as mentioned in Section \ref{section:sentence_modeling}, NMT models still struggle with some semantic concepts, which will also be a likely area of focus in years to come. While examining some of these failings of NMT can help, predicting the future of research and development in the field is nearly impossible.

New models and methods are being reported on a daily basis with far too many advancements to survey, and state-of-the-art practices becoming outdated in a matter of months. Notable recent advancements include using caching to provide networks with greater context than simply the individual sentences being translated \cite{Kuang2018}, the ability to better handle rare words \cite{Luong2014, Sennrich2015}, and the ability to translate to and from understudied languages, such as those that are polysynthetic \cite{Mager2018}. Additionally work has been conducted on the selection, sensitivity, and tuning of hyperparameters \cite{Ott2018}, denoising of data \cite{Wang2018b}, and a number of other important topics surrounding neural machine translation. Finally, a new branch of machine translation has been opened up by groundbreaking research: multilingual translation.

A fairly recent study \cite{Johnson2016} showed that a single, simple (but large) neural network could be trained to convert a number (up to at least twelve) of different languages to each other, automatically recognizing the source language and simply needing an input token to identify the output language. Furthermore, the model was found to be capable of understanding, at least somewhat, multilingual input, and of producing mixed outputs when multiple language tokens are given, sometimes even in languages related to, but not actually, those selected. This suggests that deep neural networks may be capable of learning universal representations for information, independent of language, and even more, that they might possibly be capable of learning some etymology and relationships between and among families of different languages.
\subsection{Summary of Deep Learning NLP Applications}

\label{section:summary_of_applications}

Numerous other applications of natural language processing exist including grammar correction, as seen in word processors, and author mimicking, which, given sufficient data, generates text replicating the style of a particular writer. Many of these applications are infrequently used, understudied, or not yet exposed to deep learning. However, the area of sentiment analysis should be noted, as it is becoming increasingly popular and utilizing deep learning. In large part a semantic task, it is the extraction of a writer's sentiment---their positive, negative, or neutral inclination towards some subject or idea \cite{Jurafsky2017}. Applications are varied, including product research, futures prediction, social media analysis, and classification of spam \cite{Zheng2018,Etter2018}. The current state of the art uses an ensemble including both LSTMs and CNNs \cite{Cliche2017}. 

This section has provided a number of select examples of the applied usages of deep learning in natural language processing. Countless studies have been conducted in these and similar areas, chronicling the ways in which deep learning has facilitated the successful use of natural language in a wide variety of applications. Only a minuscule fraction of such work has been referred to in this survey.

While more specific recommendations for practitioners have been discussed in some individual subsections, the current trend in state-of-the-art models in all application areas is to use pre-trained stacks of Transformer units in some configuration, whether in encoder-decoder configurations or just as encoders. Thus, self-attention which is the mainstay of Transformer has become the norm, along with cross-attention between encoder and decoder units, if decoders are present. In fact, in many recent papers, if not most, Transformers have begun to replace LSTM units that were preponderant just a few months ago. Pre-training of these large Transformer models has also become the accepted way to endow a model with generalized knowledge of language. Models such as BERT, which have been trained on corpora of billions of words, are available for download, thus providing a practitioner with a model that possesses a great amount of general knowledge of language already. A practitioner can further train it with one's own general corpora, if desired, but such training is not always necessary, considering the enormous sizes of the pre-training that downloaded models have received. To train a model to perform a certain task well, the last step a practitioner must go through is to use available downloadable task-specific corpora, or build one's own task-specific corpus. This last training step is usually supervised. It is also recommended that if several tasks are to be performed, multi-task training be used wherever possible.
\section{Conclusions}

\label{section:conclusions}

Early applications of natural language processing included a well-acclaimed but simpleminded algebra word problem solver program called STUDENT \cite{Bobrow1964}, as well as interesting but severely constrained conversational systems such as Eliza, which acted as a ``psycho-therapist" \cite{Weizenbaum1966}), and another that conversed about manipulating blocks in a microworld \cite{Winograd1971}. Nowadays, highly advanced applications of NLP are ubiquitous. These include Google's and Microsoft's machine translators, which translate more or less competently from a language to scores of other languages, as well as a number of devices which process voice commands and respond in like. The emergence of these sophisticated applications, particularly in deployed settings, acts as a testament to the impressive accomplishments that have been made in this domain over the last sixty or so years. Without a doubt, incredible progress has taken place, particularly in the last several years.

As has been shown, this recent progress has a clear causal relationship with the remarkable advances in Artificial Neural Networks. Considered an ``old" technology just a decade ago, these machine learning constructs have ushered in progress at an unprecedented rate, breaking performance records in myriad tasks in miscellaneous fields. In particular, deep neural architectures, have instilled models with higher performance in natural language tasks,  in terms of ``imperfect" metrics. Consolidating the analysis of all the models surveyed, a few general trends can be surmised. Both convolutional and recurrent specimen had contributed to the state of the art in the recent past, however, of very late, stacks of  attention-powered Transformer units as encoders and often decoders, have consistently produced superior results across the rich and varying terrain of the NLP field. These models are generally heavily pre-trained on general language knowledge in an unsupervised or supervised manner, and somewhat lightly trained on specific tasks in a supervised fashion.    Second, attention mechanisms alone, without recurrences or convolutions, seem to provide the best connections between encoders and decoders. Third, forcing networks to examine different features (by performing multiple tasks) usually improves results. Finally, while highly engineering networks usually optimizes results, there is no substitute for cultivating networks with large quantities of high quality data, although pre-training on large generic corpora seems to help immensely. Following from this final observation, it may be useful to direct more research effort toward pre-training methodologies, rather than developing highly-specialized components to squeeze the last drops of performance from complex models.

While the numerous stellar architectures being proposed each month are highly competitive, muddling the process of identifying a winning architecture, the methods of evaluation used add just as much complexity to the problem. Datasets used to evaluate new models are often generated specifically for those models and are then used only several more times, if at all, although consolidated datasets encompassing several tasks such as GLUE \cite{Wang2018c} have started to emerge.  As the features and sizes of these datasets are highly variable, this makes comparison difficult. Most subfields of NLP, as well as the field as a whole, would benefit from extensive, large-scale discussions regarding the necessary contents of such datasets, followed by the compilation of such sets. In addition to high variability in evaluation data, there are numerous metrics used to evaluate performance on each task. Oftentimes, comparing similar models is difficult due to the fact that different metrics are reported for each. Agreement on particular sets of metrics would go a long way toward ensuring clear comparisons in the field.

Furthermore, metrics are usually only reported for the best case, with few mentions of average cases and variability, or of worst cases. While it is important to understand the possible performance of new models, it is just as important to understand the standard performance. If models produce highly variable results, they may take many attempts to train to the cutting-edge levels reported. In most cases, this is undesirable, and models that can be consistently trained to relatively high levels of performance are preferable. While increasingly large numbers of randomized parameters do reduce variation in performance, some variance will always exist, necessitating the reporting of more than just best-case metrics.

One final recommendation for future work is that it be directed toward a wider variety of languages than it is at present. Currently, the vast majority of research in NLP is conducted on the English language, with another sizeable portion using Mandarin Chinese. In translation tasks, English is almost always either the input or output language, with the other end usually being one of a dozen major European or Eastern Asian languages. This neglects entire families of languages, as well as the people who speak them. Many linguistic intricacies may not be expressed in any of the languages used, and therefore are not captured in current NLP software. Furthermore, there are thousands of languages spoken throughout the world, with at least eighty spoken by more than 10 million people, meaning that current research excludes an immense segment of humankind. Collection and validation of data in under-analyzed languages, as well as testing NLP models using such data, will be a tremendous contribution to not only the field of natural language processing, but to human society as a whole.

Due to the small amounts of data available in many languages, the authors do not foresee the complete usurpation of traditional NLP models by deep learning any time in the near future. Deep learning models (and even shallow ANNs) are extremely data-hungry. Contrastingly, many traditional models require only relatively small amounts of training data. However, looking further forward, it can be anticipated that deep learning models will become the norm in computational linguistics, with pre-training and transfer learning playing highly impactful roles. Collobert et al. \cite{Collobert2011} sparked the deep learning revolution in NLP, although one of the key contributions of their work---that of a single unified model---was not realized widely. Instead, neural networks were introduced into traditional NLP tasks, and are only now reconnecting. In the field of parsing, for example, most models continue to implement non-neural structures, simply using ANNs on the side to make the decisions that were previously done using rules and probability models. While more versatile and general architectures are obviously becoming more and more of a reality, understanding the abstract concepts handled by such networks is important to understanding how to build and train better networks. Furthermore, as abstraction is a hallmark of human intelligence, understanding of the abstractions that take place inside an ANN may aid in the understanding of human intelligence and the processes that underlie it. Just as human linguistic ability is only a piece of our sentience, so is linguistic processing just a small piece of artificial intelligence. Understanding how such components are interrelated is important in constructing more complete AI systems, and creating a unified NLP architecture is another step toward making such a system a reality.

This goal will also be aided by further advances in computational equipment. While GPUs have significantly improved the ability to train deep networks, they are only a step in the right direction \cite{Schuman2017}. The next step is the wider availability of chips designed specifically for this purpose, such as Google's Tensor Processing Unit (TPU), Microsoft's Catapult, and Intel's Lake Crest \cite{Hennessy2017}. Ultimately, artificial neural networks implemented in traditional von Neumann style computers may not be able to reach their full potential. Luckily, another old line of work in computer science and engineering has seen a resurgance in recent years: neuromorphic computing. With neuromorphic chips, which implement neural structures at the hardware level, expected much more widely in coming years \cite{Monroe2014}, the continuation of deep learning and the longevity of its success can be highly anticipated, ensuring the opportunity for sustained progress in natural language processing.

\balance

\bibliographystyle{IEEEtran}
\bibliography{references}

\begin{thebibliography}{100}
\providecommand{\url}[1]{#1}
\csname url@samestyle\endcsname
\providecommand{\newblock}{\relax}
\providecommand{\bibinfo}[2]{#2}
\providecommand{\BIBentrySTDinterwordspacing}{\spaceskip=0pt\relax}
\providecommand{\BIBentryALTinterwordstretchfactor}{4}
\providecommand{\BIBentryALTinterwordspacing}{\spaceskip=\fontdimen2\font plus
\BIBentryALTinterwordstretchfactor\fontdimen3\font minus
  \fontdimen4\font\relax}
\providecommand{\BIBforeignlanguage}[2]{{%
\expandafter\ifx\csname l@#1\endcsname\relax
\typeout{** WARNING: IEEEtran.bst: No hyphenation pattern has been}%
\typeout{** loaded for the language `#1'. Using the pattern for}%
\typeout{** the default language instead.}%
\else
\language=\csname l@#1\endcsname
\fi
#2}}
\providecommand{\BIBdecl}{\relax}
\BIBdecl

\bibitem{Jones1994}
K.~S. Jones, ``Natural language processing: a historical review,'' in
  \emph{Current Issues in Computational Linguistics: in Honour of Don
  Walker}.\hskip 1em plus 0.5em minus 0.4em\relax Springer, 1994, pp. 3--16.

\bibitem{Liddy2001}
E.~D. Liddy, ``Natural language processing,'' 2001.

\bibitem{Coates2013}
A.~Coates, B.~Huval, T.~Wang, D.~Wu, B.~Catanzaro, and N.~Andrew, ``Deep
  learning with cots hpc systems,'' in \emph{ICML}, 2013, pp. 1337--1345.

\bibitem{Raina2009}
R.~Raina, A.~Madhavan, and A.~Y. Ng, ``Large-scale deep unsupervised learning
  using graphics processors,'' in \emph{ICML}, 2009, pp. 873--880.

\bibitem{Goodfellow2016}
I.~Goodfellow, Y.~Bengio, A.~Courville, and Y.~Bengio, \emph{Deep
  learning}.\hskip 1em plus 0.5em minus 0.4em\relax MIT Press, Cambridge, 2016,
  vol.~1.

\bibitem{LeCun2015}
Y.~LeCun, Y.~Bengio, and G.~Hinton, ``Deep learning,'' \emph{Nature}, vol. 521,
  no. 7553, pp. 436--444, 2015.

\bibitem{Schmidhuber2015}
J.~Schmidhuber, ``Deep learning in neural networks: An overview,'' \emph{Neural
  networks}, vol.~61, pp. 85--117, 2015.

\bibitem{Ciresan2011}
D.~Ciresan, U.~Meier, J.~Masci, L.~Maria~Gambardella, and J.~Schmidhuber,
  ``Flexible, high performance convolutional neural networks for image
  classification,'' in \emph{IJCAI}, vol.~22, no.~1, 2011, p. 1237.

\bibitem{Collobert2011}
R.~Collobert, J.~Weston, L.~Bottou, M.~Karlen, K.~Kavukcuoglu, and P.~Kuksa,
  ``Natural language processing (almost) from scratch,'' \emph{Journal of
  Machine Learning Research}, vol.~12, pp. 2493--2537, 2011.

\bibitem{Goldberg2017}
Y.~Goldberg, ``Neural network methods for natural language processing,''
  \emph{Synthesis Lectures on Human Language Technologies}, vol.~10, no.~1, pp.
  1--309, 2017.

\bibitem{Liu2018a}
Y.~Liu and M.~Zhang, ``Neural network methods for natural language
  processing,'' 2018.

\bibitem{Young2018}
T.~Young, D.~Hazarika, S.~Poria, and E.~Cambria, ``Recent trends in deep
  learning based natural language processing,'' \emph{ieee Computational
  intelligenCe magazine}, vol.~13, no.~3, pp. 55--75, 2018.

\bibitem{Rumelhart1985}
D.~Rumelhart, G.~Hinton, and R.~Williams, ``Learning internal representations
  by error propagation,'' UCSD, Tech. Rep., 1985.

\bibitem{Lecun1989}
Y.~LeCun, B.~Boser, J.~S. Denker, D.~Henderson, R.~E. Howard, W.~Hubbard, and
  L.~D. Jackel, ``Backpropagation applied to handwritten zip code
  recognition,'' \emph{Neural Computation}, vol.~1, no.~4, 1989.

\bibitem{LeCun1998}
Y.~LeCun, L.~Bottou, Y.~Bengio, and P.~Haffner, ``Gradient-based learning
  applied to document recognition,'' \emph{Proc of the IEEE}, vol.~86, no.~11,
  pp. 2278--2324, 1998.

\bibitem{Fukushima1980}
K.~Fukushima, ``Neocognitron: A self-organizing neural network model for a
  mechanism of pattern recognition unaffected by shift in position,''
  \emph{Bioological Cybernetics}, vol.~36, pp. 193--202, 1980.

\bibitem{Fukushima1982}
K.~Fukushima and S.~Miyake, ``Neocognitron: A new algorithm for pattern
  recognition tolerant of deformations and shifts in position,'' \emph{Pattern
  Recognition}, vol.~15, no.~6, pp. 455--469, 1982.

\bibitem{LeCun1995}
Y.~LeCun, Y.~Bengio \emph{et~al.}, ``Convolutional networks for images, speech,
  and time series,'' \emph{The Handbook of Brain Theory and Neural Networks},
  vol. 3361, no.~10, 1995.

\bibitem{Krizhevsky2014}
A.~Krizhevsky, ``One weird trick for parallelizing convolutional neural
  networks,'' \emph{arXiv preprint arXiv:1404.5997}, 2014.

\bibitem{Kim2014}
Y.~Kim, ``Convolutional neural networks for sentence classification,''
  \emph{arXiv preprint arXiv:1408.5882}, 2014.

\bibitem{Kalchbrenner2014}
N.~Kalchbrenner, E.~Grefenstette, and P.~Blunsom, ``A convolutional neural
  network for modelling sentences,'' \emph{arXiv preprint arXiv:1404.2188},
  2014.

\bibitem{DosSantos2014}
C.~N. Dos~Santos and M.~Gatti, ``Deep convolutional neural networks for
  sentiment analysis of short texts.'' in \emph{COLING}, 2014, pp. 69--78.

\bibitem{Zeng2014}
D.~Zeng, K.~Liu, S.~Lai, G.~Zhou, J.~Zhao \emph{et~al.}, ``Relation
  classification via convolutional deep neural network.'' in \emph{COLING},
  2014.

\bibitem{Kawato1987}
M.~Kawato, K.~Furukawa, and R.~Suzuki, ``A hierarchical neural-network model
  for control and learning of voluntary movement,'' \emph{Biological
  Cybernetics}, vol.~57, no.~3, pp. 169--185, 1987.

\bibitem{Goller1996}
C.~Goller and A.~Kuchler, ``Learning task-dependent distributed representations
  by backpropagation through structure,'' in \emph{IEEE International Conf on
  Neural Networks}, vol.~1, 1996, pp. 347--352.

\bibitem{Socher2011a}
R.~Socher, E.~Huang, J.~Pennin, C.~Manning, and A.~Ng, ``Dynamic pooling and
  unfolding recursive autoencoders for paraphrase detection,'' in \emph{NIPS},
  2011, pp. 801--809.

\bibitem{Elman1990}
J.~L. Elman, ``Finding structure in time,'' \emph{Cognitive Science}, vol.~14,
  no.~2, pp. 179--211, 1990.

\bibitem{Fausett1994}
L.~Fausett, \emph{Fundamentals of neural networks: architectures, algorithms,
  and applications}.\hskip 1em plus 0.5em minus 0.4em\relax Prentice-Hall,
  Inc., 1994.

\bibitem{Mikolov2010}
T.~Mikolov, M.~Karafi{\'a}t, L.~Burget, J.~{\v{C}}ernock{\`y}, and
  S.~Khudanpur, ``Recurrent neural network based language model,'' in
  \emph{Annual Conf of the Intnl Speech Communication Assoc}, vol.~2, 2010,
  p.~3.

\bibitem{Mikolov2011a}
T.~Mikolov, S.~Kombrink, L.~Burget, J.~{\v{C}}ernock{\`y}, and S.~Khudanpur,
  ``Extensions of recurrent neural network language model,'' in \emph{IEEE
  ICASSP}, 2011, pp. 5528--5531.

\bibitem{Mikolov2011b}
T.~Mikolov, A.~Deoras, D.~Povey, L.~Burget, and J.~{\v{C}}ernock{\`y},
  ``Strategies for training large scale neural network language models,'' in
  \emph{IEEE Workshop on Automatic Speech Recognition and Understanding}, 2011.

\bibitem{Schmidhuber1992}
J.~Schmidhuber, ``Learning complex, extended sequences using the principle of
  history compression,'' \emph{Neural Computation}, vol.~4, no.~2, pp.
  234--242, 1992.

\bibitem{ElHihi1996}
S.~El~Hihi and Y.~Bengio, ``Hierarchical recurrent neural networks for
  long-term dependencies,'' in \emph{NIPS}, 1996, pp. 493--499.

\bibitem{Hochreiter1997}
S.~Hochreiter and J.~Schmidhuber, ``Long short-term memory,'' \emph{Neural
  computation}, vol.~9, no.~8, pp. 1735--1780, 1997.

\bibitem{Greff2017}
K.~Greff, R.~K. Srivastava, J.~Koutn{\'\i}k, B.~R. Steunebrink, and
  J.~Schmidhuber, ``Lstm: A search space odyssey,'' \emph{IEEE Transactions on
  Neural Networks and Learning Systems}, vol.~28, no.~10, 2017.

\bibitem{Cho2014a}
K.~Cho, B.~Van~Merri{\"e}nboer, D.~Bahdanau, and Y.~Bengio, ``On the properties
  of neural machine translation: Encoder-decoder approaches,'' \emph{arXiv
  preprint arXiv:1409.1259}, 2014.

\bibitem{Chung2014}
J.~Chung, C.~Gulcehre, K.~Cho, and Y.~Bengio, ``Empirical evaluation of gated
  recurrent neural networks on sequence modeling,'' \emph{arXiv preprint
  arXiv:1412.3555}, 2014.

\bibitem{Bahdanau2014}
D.~Bahdanau, K.~Cho, and Y.~Bengio, ``Neural machine translation by jointly
  learning to align and translate,'' \emph{arXiv preprint arXiv:1409.0473},
  2014.

\bibitem{Rush2015}
A.~Rush, S.~Chopra, and J.~Weston, ``A neural attention model for abstractive
  sentence summarization,'' \emph{arXiv preprint arXiv:1509.00685}, 2015.

\bibitem{Paulus2017}
R.~Paulus, C.~Xiong, and R.~Socher, ``A deep reinforced model for abstractive
  summarization,'' \emph{arXiv preprint arXiv:1705.04304}, 2017.

\bibitem{Wang2017}
W.~Wang, N.~Yang, F.~Wei, B.~Chang, and M.~Zhou, ``Gated self-matching networks
  for reading comprehension and question answering,'' in \emph{ACL}, vol.~1,
  2017, pp. 189--198.

\bibitem{Vaswani2017}
A.~Vaswani, N.~Shazeer, N.~Parmar, J.~Uszkoreit, L.~Jones, A.~N. Gomez,
  {\L}.~Kaiser, and I.~Polosukhin, ``Attention is all you need,'' in
  \emph{NIPS}, 2017, pp. 6000--6010.

\bibitem{Bengio1994}
Y.~Bengio, P.~Simard, and P.~Frasconi, ``Learning long-term dependencies with
  gradient descent is difficult,'' \emph{{IEEE Transactions on Neural
  Networks}}, vol.~5, no.~2, 1994.

\bibitem{Nair2010}
V.~Nair and G.~E. Hinton, ``Rectified linear units improve restricted boltzmann
  machines,'' in \emph{ICML}, 2010, pp. 807--814.

\bibitem{He2016b}
K.~He, X.~Zhang, S.~Ren, and J.~Sun, ``Deep residual learning for image
  recognition,'' in \emph{IEEE CVPR}, 2016, pp. 770--778.

\bibitem{Srivastava2015}
R.~K. Srivastava, K.~Greff, and J.~Schmidhuber, ``Highway networks,''
  \emph{arXiv preprint arXiv:1505.00387}, 2015.

\bibitem{Huang2017}
G.~Huang, Z.~Liu, K.~Q. Weinberger, and L.~van~der Maaten, ``Densely connected
  convolutional networks,'' in \emph{IEEE CVPR}, vol.~1, no.~2, 2017.

\bibitem{Jurafsky2008}
D.~Jurafsky and J.~Martin, \emph{Speech \& language processing}.\hskip 1em plus
  0.5em minus 0.4em\relax Pearson Education, 2000.

\bibitem{Bengio2003}
Y.~Bengio, R.~Ducharme, P.~Vincent, and C.~Jauvin, ``A neural probabilistic
  language model,'' \emph{{J. of Machine Learning Research}}, vol.~3, 2003.

\bibitem{DeMulder2015}
W.~De~Mulder, S.~Bethard, and M.-F. Moens, ``A survey on the application of
  recurrent neural networks to statistical language modeling,'' \emph{Computer
  Speech \& Language}, vol.~30, no.~1, pp. 61--98, 2015.

\bibitem{Iyer1997}
R.~Iyer, M.~Ostendorf, and M.~Meteer, ``Analyzing and predicting language model
  improvements,'' in \emph{IEEE Workshop on Automatic Speech Recognition and
  Understanding}, 1997, pp. 254--261.

\bibitem{Chen1998}
S.~F. Chen, D.~Beeferman, and R.~Rosenfeld, ``Evaluation metrics for language
  models,'' 1998.

\bibitem{Clarkson2001}
P.~Clarkson and T.~Robinson, ``Improved language modelling through better
  language model evaluation measures,'' \emph{Computer Speech \& Language},
  vol.~15, no.~1, pp. 39--53, 2001.

\bibitem{Marcus1993}
M.~Marcus, M.~A. Marcinkiewicz, and B.~Santorini, ``Building a large annotated
  corpus of english: The penn treebank,'' \emph{Computational Linguistics},
  vol.~19, no.~2, pp. 313--330, 1993.

\bibitem{Chelba2013}
C.~Chelba, T.~Mikolov, M.~Schuster, Q.~Ge, T.~Brants, P.~Koehn, and
  T.~Robinson, ``One billion word benchmark for measuring progress in
  statistical language modeling,'' \emph{arXiv preprint arXiv:1312.3005}, 2013.

\bibitem{Daniluk2017}
M.~Daniluk, T.~Rockt{\"a}schel, J.~Welbl, and S.~Riedel, ``Frustratingly short
  attention spans in neural language modeling,'' \emph{arXiv preprint
  arXiv:1702.04521}, 2017.

\bibitem{Benes2017}
K.~Bene{\v{s}}, M.~K. Baskar, and L.~Burget, ``Residual memory networks in
  language modeling: Improving the reputation of feed-forward networks,''
  \emph{Interspeech 2017}, pp. 284--288, 2017.

\bibitem{Pham2016}
N.-Q. Pham, G.~Kruszewski, and G.~Boleda, ``Convolutional neural network
  language models,'' in \emph{EMNLP}, 2016, pp. 1153--1162.

\bibitem{Lin2013}
M.~Lin, Q.~Chen, and S.~Yan, ``Network in network,'' \emph{arXiv preprint
  arXiv:1312.4400}, 2013.

\bibitem{Kim2016}
Y.~Kim, Y.~Jernite, D.~Sontag, and A.~M. Rush, ``Character-aware neural
  language models.'' in \emph{AAAI}, 2016, pp. 2741--2749.

\bibitem{Botha2014}
J.~Botha and P.~Blunsom, ``Compositional morphology for word representations
  and language modelling,'' in \emph{ICML}, 2014, pp. 1899--1907.

\bibitem{Zaremba2014}
W.~Zaremba, I.~Sutskever, and O.~Vinyals, ``Recurrent neural network
  regularization,'' \emph{arXiv preprint arXiv:1409.2329}, 2014.

\bibitem{Jozefowicz2016}
R.~Jozefowicz, O.~Vinyals, M.~Schuster, N.~Shazeer, and Y.~Wu, ``Exploring the
  limits of language modeling,'' 2016.

\bibitem{Ji2015}
Y.~Ji, T.~Cohn, L.~Kong, C.~Dyer, and J.~Eisenstein, ``Document context
  language models,'' \emph{arXiv preprint arXiv:1511.03962}, 2015.

\bibitem{Shazeer2015}
N.~Shazeer, J.~Pelemans, and C.~Chelba, ``Sparse non-negative matrix language
  modeling for skip-grams,'' in \emph{Interspeech 2015}, 2015.

\bibitem{Williams2015}
W.~Williams, N.~Prasad, D.~Mrva, T.~Ash, and T.~Robinson, ``Scaling recurrent
  neural network language models,'' in \emph{IEEE ICASSP}, 2015.

\bibitem{Mikolov2013a}
T.~Mikolov, K.~Chen, G.~Corrado, and J.~Dean, ``Efficient estimation of word
  representations in vector space,'' \emph{arXiv preprint arXiv:1301.3781},
  2013.

\bibitem{Mikolov2013b}
T.~Mikolov, I.~Sutskever, K.~Chen, G.~S. Corrado, and J.~Dean, ``Distributed
  representations of words and phrases and their compositionality,'' in
  \emph{NIPS}, 2013, pp. 3111--3119.

\bibitem{Radford2018}
A.~Radford, K.~Narasimhan, T.~Salimans, and I.~Sutskever, ``Improving language
  understanding by generative pre-training,'' \emph{URL
  https://s3-us-west-2.amazonaws.com/openai-assets/research-covers/language-unsupervised/language\_understanding\_paper.pdf},
  2018.

\bibitem{Peters2018}
M.~E. Peters, M.~Neumann, M.~Iyyer, M.~Gardner, C.~Clark, K.~Lee, and
  L.~Zettlemoyer, ``Deep contextualized word representations,'' \emph{arXiv
  preprint arXiv:1802.05365}, 2018.

\bibitem{Devlin2018}
J.~Devlin, M.-W. Chang, K.~Lee, and K.~Toutanova, ``Bert: Pre-training of deep
  bidirectional transformers for language understanding,'' \emph{arXiv preprint
  arXiv:1810.04805}, 2018.

\bibitem{Liu2019b}
X.~Liu, P.~He, W.~Chen, and J.~Gao, ``Multi-task deep neural networks for
  natural language understanding,'' \emph{arXiv preprint arXiv:1901.11504},
  2019.

\bibitem{Liu2017}
X.~Liu, Y.~Shen, K.~Duh, and J.~Gao, ``Stochastic answer networks for machine
  reading comprehension,'' \emph{arXiv preprint arXiv:1712.03556}, 2017.

\bibitem{Liu2018b}
X.~Liu, K.~Duh, and J.~Gao, ``Stochastic answer networks for natural language
  inference,'' \emph{arXiv preprint arXiv:1804.07888}, 2018.

\bibitem{McCoy2019}
R.~T. McCoy, E.~Pavlick, and T.~Linzen, ``Right for the wrong reasons:
  Diagnosing syntactic heuristics in natural language inference,'' \emph{ACL},
  2019.

\bibitem{Luong2013}
T.~Luong, R.~Socher, and C.~Manning, ``Better word representations with
  recursive neural networks for morphology,'' in \emph{CoNLL}, 2013.

\bibitem{Finkelstein2001}
L.~Finkelstein, E.~Gabrilovich, Y.~Matias, E.~Rivlin, Z.~Solan, G.~Wolfman, and
  E.~Ruppin, ``Placing search in context: The concept revisited,'' in
  \emph{Intnl Conf on World Wide Web}, 2001, pp. 406--414.

\bibitem{Creutz2007}
M.~Creutz and K.~Lagus, ``Unsupervised models for morpheme segmentation and
  morphology learning,'' \emph{ACM TSLP}, vol.~4, no.~1, 2007.

\bibitem{Miller1991}
G.~A. Miller and W.~G. Charles, ``Contextual correlates of semantic
  similarity,'' \emph{Language and Cognitive Processes}, vol.~6, no.~1, 1991.

\bibitem{Rubenstein1965}
H.~Rubenstein and J.~B. Goodenough, ``Contextual correlates of synonymy,''
  \emph{CACM}, vol.~8, no.~10, pp. 627--633, 1965.

\bibitem{Huang2012}
R.~Huang, Eric Hand~Socher, C.~Manning, and A.~Ng, ``Improving word
  representations via global context and multiple word prototypes,'' in
  \emph{ACL: Vol 1}, 2012, pp. 873--882.

\bibitem{Belinkov2017}
Y.~Belinkov, N.~Durrani, F.~Dalvi, H.~Sajjad, and J.~Glass, ``What do neural
  machine translation models learn about morphology?'' \emph{arXiv preprint
  arXiv:1704.03471}, 2017.

\bibitem{Cettolo2012}
M.~Cettolo, C.~Girardi, and M.~Federico, ``Wit3: Web inventory of transcribed
  and translated talks,'' in \emph{Conf of European Assoc. for Machine
  Translation}, 2012, pp. 261--268.

\bibitem{Cettolo2016}
M.~Cettolo, ``An arabic-hebrew parallel corpus of ted talks,'' \emph{arXiv
  preprint arXiv:1610.00572}, 2016.

\bibitem{Morita2015}
H.~Morita, D.~Kawahara, and S.~Kurohashi, ``Morphological analysis for
  unsegmented languages using recurrent neural network language model,'' in
  \emph{EMNLP}, 2015, pp. 2292--2297.

\bibitem{Kawahara2006}
D.~Kawahara and S.~Kurohashi, ``Case frame compilation from the web using
  high-performance computing,'' in \emph{LREC}, 2006, pp. 1344--1347.

\bibitem{Kawahara2002}
D.~Kawahara, S.~Kurohashi, and K.~Hasida, ``Construction of a japanese
  relevance-tagged corpus.'' in \emph{LREC}, 2002.

\bibitem{Hangyo2012}
M.~Hangyo, D.~Kawahara, and S.~Kurohashi, ``Building a diverse document leads
  corpus annotated with semantic relations,'' in \emph{Pacific-Asia Conf on
  Language, Information, \& Computation}, 2012, pp. 535--544.

\bibitem{Dehouck2018}
M.~Dehouck and P.~Denis, ``A framework for understanding the role of morphology
  in universal dependency parsing,'' 2018.

\bibitem{More2018}
A.~More, {\"O}.~{\c{C}}etino{\u{g}}lu, {\c{C}}.~{\c{C}}{\"o}ltekin, N.~Habash,
  B.~Sagot, D.~Seddah, D.~Taji, and R.~Tsarfaty, ``Conll-ul: Universal
  morphological lattices for universal dependency parsing,'' in
  \emph{Proceedings of the Eleventh International Conference on Language
  Resources and Evaluation}, 2018.

\bibitem{Nivre2003}
J.~Nivre, ``An efficient algorithm for projective dependency parsing,'' in
  \emph{Intnl Workshop on Parsing Technologies}, 2003.

\bibitem{Nivre2004}
------, ``Incrementality in deterministic dependency parsing,'' in
  \emph{Workshop on Incremental Parsing: Bringing Engineering and Cognition
  Together}, 2004, pp. 50--57.

\bibitem{Nivre2009}
J.~Nivre, M.~Kuhlmann, and J.~Hall, ``An improved oracle for dependency parsing
  with online reordering,'' in \emph{Intnl Conf on Parsing Technologies}, 2009,
  pp. 73--76.

\bibitem{Socher2013a}
R.~Socher, A.~Perelygin, J.~Wu, J.~Chuang, C.~D. Manning, A.~Ng, and C.~Potts,
  ``Recursive deep models for semantic compositionality over a sentiment
  treebank,'' in \emph{EMNLP}, 2013, pp. 1631--1642.

\bibitem{Socher2013b}
R.~Socher, J.~Bauer, C.~Manning \emph{et~al.}, ``Parsing with compositional
  vector grammars,'' in \emph{ACL}, vol.~1, 2013, pp. 455--465.

\bibitem{Fujisaki1991}
T.~Fujisaki, F.~Jelinek, J.~Cocke, E.~Black, and T.~Nishino, ``A probabilistic
  parsing method for sentence disambiguation,'' in \emph{Current issues in
  Parsing Technology}, 1991, pp. 139--152.

\bibitem{Jelinek1992}
F.~Jelinek, J.~Lafferty, and R.~Mercer, ``Basic methods of probabilistic
  context free grammars,'' in \emph{Speech Recognition and
  Understanding}.\hskip 1em plus 0.5em minus 0.4em\relax Springer, 1992, pp.
  345--360.

\bibitem{Le2014a}
P.~Le and W.~Zuidema, ``The inside-outside recursive neural network model for
  dependency parsing,'' in \emph{EMNLP}, 2014, pp. 729--739.

\bibitem{Vinyals2015a}
O.~Vinyals, {\L}.~Kaiser, T.~Koo, S.~Petrov, I.~Sutskever, and G.~Hinton,
  ``Grammar as a foreign language,'' in \emph{NIPS}, 2015, pp. 2773--2781.

\bibitem{Petrov2012}
S.~Petrov and R.~McDonald, ``Overview of the 2012 shared task on parsing the
  web,'' in \emph{Notes of the 1st Workshop on Syntactic Analysis of
  Non-canonical Language}, vol.~59, 2012.

\bibitem{Judge2006}
J.~Judge, A.~Cahill, and J.~Van~Genabith, ``Questionbank: Creating a corpus of
  parse-annotated questions,'' in \emph{COLING}, 2006, pp. 497--504.

\bibitem{Stenetorp2013}
P.~Stenetorp, ``Transition-based dependency parsing using recursive neural
  networks,'' in \emph{NIPS Workshop on Deep Learning}, 2013.

\bibitem{Surdeanu2008}
M.~Surdeanu, R.~Johansson, A.~Meyers, L.~M{\`a}rquez, and J.~Nivre, ``The
  conll-2008 shared task on joint parsing of syntactic and semantic
  dependencies,'' in \emph{CONLL}, 2008, pp. 159--177.

\bibitem{Chen2014}
D.~Chen and C.~Manning, ``A fast and accurate dependency parser using neural
  networks,'' in \emph{EMNLP}, 2014, pp. 740--750.

\bibitem{Zhou2015}
H.~Zhou, Y.~Zhang, S.~Huang, and J.~Chen, ``A neural probabilistic
  structured-prediction model for transition-based dependency parsing,'' in
  \emph{ACL and IJCNLP}, vol.~1, 2015, pp. 1213--1222.

\bibitem{Weiss2015}
D.~Weiss, C.~Alberti, M.~Collins, and S.~Petrov, ``Structured training for
  neural network transition-based parsing,'' \emph{arXiv preprint
  arXiv:1506.06158}, 2015.

\bibitem{Li2014}
Z.~Li, M.~Zhang, and W.~Chen, ``Ambiguity-aware ensemble training for
  semi-supervised dependency parsing,'' in \emph{ACL}, vol.~1, 2014.

\bibitem{Dyer2015}
C.~Dyer, M.~Ballesteros, W.~Ling, A.~Matthews, and N.~A. Smith,
  ``Transition-based dependency parsing with stack long short-term memory,''
  \emph{arXiv preprint arXiv:1505.08075}, 2015.

\bibitem{deMarneffe2008}
M.-C. De~Marneffe and C.~D. Manning, ``The stanford typed dependencies
  representation,'' in \emph{COLING Workshop on Cross-framework and
  Cross-domain Parser Evaluation}, 2008, pp. 1--8.

\bibitem{Xue2005}
N.~Xue, F.~Xia, F.-D. Chiou, and M.~Palmer, ``The penn chinese treebank: Phrase
  structure annotation of a large corpus,'' \emph{Natural Language
  Engineering}, vol.~11, no.~2, pp. 207--238, 2005.

\bibitem{Andor2016}
D.~Andor, C.~Alberti, D.~Weiss, A.~Severyn, A.~Presta, K.~Ganchev, S.~Petrov,
  and M.~Collins, ``Globally normalized transition-based neural networks,''
  \emph{arXiv preprint arXiv:1603.06042}, 2016.

\bibitem{Wang2018a}
Y.~Wang, W.~Che, J.~Guo, and T.~Liu, ``A neural transition-based approach for
  semantic dependency graph parsing,'' 2018.

\bibitem{Cross2016}
J.~Cross and L.~Huang, ``Incremental parsing with minimal features using
  bi-directional lstm,'' \emph{arXiv preprint arXiv:1606.06406}, 2016.

\bibitem{Tai2015}
K.~S. Tai, R.~Socher, and C.~D. Manning, ``Improved semantic representations
  from tree-structured long short-term memory networks,'' \emph{arXiv preprint
  arXiv:1503.00075}, 2015.

\bibitem{Oepen2015}
S.~Oepen, M.~Kuhlmann, Y.~Miyao, D.~Zeman, S.~Cinkov{\'a}, D.~Flickinger,
  J.~Hajic, and Z.~Uresova, ``Semeval 2015 task 18: Broad-coverage semantic
  dependency parsing,'' in \emph{Intnl Workshop on Semantic Evaluation}, 2015,
  pp. 915--926.

\bibitem{Che2012}
W.~Che, M.~Zhang, Y.~Shao, and T.~Liu, ``Semeval-2012 task 9: Chinese semantic
  dependency parsing,'' in \emph{Conference on Lexical and Computational
  Semantics}, 2012, pp. 378--384.

\bibitem{Yih2014}
W.-t. Yih, X.~He, and C.~Meek, ``Semantic parsing for single-relation question
  answering,'' in \emph{Proceedings of the 52nd Annual Meeting of the ACL
  (Volume 2: Short Papers)}, 2014, pp. 643--648.

\bibitem{Krishnamurthy2017}
J.~Krishnamurthy, P.~Dasigi, and M.~Gardner, ``Neural semantic parsing with
  type constraints for semi-structured tables,'' in \emph{Proceedings of the
  2017 Conference on Empirical Methods in NLP}, 2017, pp. 1516--1526.

\bibitem{Dyer2016}
C.~Dyer, A.~Kuncoro, M.~Ballesteros, and N.~A. Smith, ``Recurrent neural
  network grammars,'' \emph{arXiv preprint arXiv:1602.07776}, 2016.

\bibitem{Do2016}
D.~K. Choe and E.~Charniak, ``Parsing as language modeling,'' in \emph{EMNLP},
  2016, pp. 2331--2336.

\bibitem{Fried2017}
D.~Fried, M.~Stern, and D.~Klein, ``Improving neural parsing by disentangling
  model combination and reranking effects,'' \emph{arXiv preprint
  arXiv:1707.03058}, 2017.

\bibitem{Dozat2018}
T.~Dozat and C.~D. Manning, ``Simpler but more accurate semantic dependency
  parsing,'' \emph{arXiv preprint arXiv:1807.01396}, 2018.

\bibitem{Tan2018}
Z.~Tan, M.~Wang, J.~Xie, Y.~Chen, and X.~Shi, ``Deep semantic role labeling
  with self-attention,'' in \emph{Thirty-Second AAAI Conference on Artificial
  Intelligence}, 2018.

\bibitem{Duong2018}
L.~Duong, H.~Afshar, D.~Estival, G.~Pink, P.~Cohen, and M.~Johnson, ``Active
  learning for deep semantic parsing,'' in \emph{Proceedings of the 56th Annual
  Meeting of the ACL (Vol. 2: Short Papers)}, 2018, pp. 43--48.

\bibitem{Nivre2015}
J.~Nivre, ``Towards a universal grammar for natural language processing,'' in
  \emph{International Conference on Intelligent Text Processing and
  Computational Linguistics}.\hskip 1em plus 0.5em minus 0.4em\relax Springer,
  2015, pp. 3--16.

\bibitem{Zeman2018}
D.~Zeman, J.~Haji{\v{c}}, M.~Popel, M.~Potthast, M.~Straka, F.~Ginter,
  J.~Nivre, and S.~Petrov, ``Conll 2018 shared task: multilingual parsing from
  raw text to universal dependencies,'' 2018.

\bibitem{Hershcovich2018}
D.~Hershcovich, O.~Abend, and A.~Rappoport, ``Universal dependency parsing with
  a general transition-based dag parser,'' \emph{arXiv preprint
  arXiv:1808.09354}, 2018.

\bibitem{Ji2018}
T.~Ji, Y.~Liu, Y.~Wang, Y.~Wu, and M.~Lan, ``Antnlp at conll 2018 shared task:
  A graph-based parser for universal dependency parsing,'' in \emph{Proceedings
  of the CoNLL 2018 Shared Task: Multilingual Parsing from Raw Text to
  Universal Dependencies}, 2018, pp. 248--255.

\bibitem{Qi2019}
P.~Qi, T.~Dozat, Y.~Zhang, and C.~D. Manning, ``Universal dependency parsing
  from scratch,'' \emph{arXiv preprint arXiv:1901.10457}, 2019.

\bibitem{Liu2018c}
Y.~Liu, Y.~Zhu, W.~Che, B.~Qin, N.~Schneider, and N.~A. Smith, ``Parsing tweets
  into universal dependencies,'' \emph{arXiv preprint arXiv:1804.08228}, 2018.

\bibitem{Pennington2014}
J.~Pennington, R.~Socher, and C.~Manning, ``Glove: Global vectors for word
  representation,'' in \emph{EMNLP}, 2014, pp. 1532--1543.

\bibitem{Harris1954}
Z.~S. Harris, ``Distributional structure,'' \emph{Word}, vol.~10, no. 2-3, pp.
  146--162, 1954.

\bibitem{Hu2014a}
B.~Hu, Z.~Lu, H.~Li, and Q.~Chen, ``Convolutional neural network architectures
  for matching natural language sentences,'' in \emph{NIPS}, 2014, pp.
  2042--2050.

\bibitem{Bordes2014}
A.~Bordes, X.~Glorot, J.~Weston, and Y.~Bengio, ``A semantic matching energy
  function for learning with multi-relational data,'' \emph{Machine Learning},
  vol.~94, no.~2, pp. 233--259, 2014.

\bibitem{Yin2015}
W.~Yin and H.~Sch{\"u}tze, ``Convolutional neural network for paraphrase
  identification,'' in \emph{NAACL: HLT}, 2015, pp. 901--911.

\bibitem{Dolan2004}
B.~Dolan, C.~Quirk, and C.~Brockett, ``Unsupervised construction of large
  paraphrase corpora: Exploiting massively parallel news sources,'' in
  \emph{COLING}, 2004, p. 350.

\bibitem{He2015}
H.~He, K.~Gimpel, and J.~Lin, ``Multi-perspective sentence similarity modeling
  with convolutional neural networks,'' in \emph{EMNLP}, 2015.

\bibitem{Marelli2014}
M.~Marelli, L.~Bentivogli, M.~Baroni, R.~Bernardi, S.~Menini, and
  R.~Zamparelli, ``Semeval-2014 task 1: Evaluation of compositional
  distributional semantic models on full sentences through semantic relatedness
  and textual entailment,'' in \emph{Intnl Workshop on Semantic Evaluation},
  2014, pp. 1--8.

\bibitem{Agirre2012}
E.~Agirre, M.~Diab, D.~Cer, and A.~Gonzalez-Agirre, ``Semeval-2012 task 6: A
  pilot on semantic textual similarity,'' in \emph{Joint Conf on Lexical and
  Computational Semantics-Vol 1}, 2012, pp. 385--393.

\bibitem{He2016a}
H.~He and J.~Lin, ``Pairwise word interaction modeling with deep neural
  networks for semantic similarity measurement,'' in \emph{NAACL: HLT}, 2016,
  pp. 937--948.

\bibitem{Agirre2014}
E.~Agirre, C.~Banea, C.~Cardie, D.~Cer, M.~Diab, A.~Gonzalez-Agirre, W.~Guo,
  R.~Mihalcea, G.~Rigau, and J.~Wiebe, ``Semeval-2014 task 10: Multilingual
  semantic textual similarity,'' in \emph{Intnl Workshop on Semantic
  Evaluation}, 2014, pp. 81--91.

\bibitem{Yang2015}
Y.~Yang, W.-t. Yih, and C.~Meek, ``Wikiqa: A challenge dataset for open-domain
  question answering,'' in \emph{EMNLP}, 2015, pp. 2013--2018.

\bibitem{Wang2007}
M.~Wang, N.~A. Smith, and T.~Mitamura, ``What is the jeopardy model? a
  quasi-synchronous grammar for qa,'' in \emph{Joint EMNLP and CoNLL}, 2007.

\bibitem{Le2014b}
Q.~Le and T.~Mikolov, ``Distributed representations of sentences and
  documents,'' in \emph{ICML}, 2014, pp. 1188--1196.

\bibitem{Go2009}
A.~Go, R.~Bhayani, and L.~Huang, ``Twitter sentiment classification using
  distant supervision,'' \emph{CS224N Project Report, Stanford}, vol.~1,
  no.~12, 2009.

\bibitem{Li2002}
X.~Li and D.~Roth, ``Learning question classifiers,'' in \emph{COLING-Vol 1},
  2002, pp. 1--7.

\bibitem{Poliak2018a}
A.~Poliak, Y.~Belinkov, J.~Glass, and B.~Van~Durme, ``On the evaluation of
  semantic phenomena in neural machine translation using natural language
  inference,'' \emph{arXiv preprint arXiv:1804.09779}, 2018.

\bibitem{Williams2017}
A.~Williams, N.~Nangia, and S.~Bowman, ``A broad-coverage challenge corpus for
  sentence understanding through inference,'' \emph{arXiv preprint
  arXiv:1704.05426}, 2017.

\bibitem{Nangia2017}
N.~Nangia, A.~Williams, A.~Lazaridou, and S.~R. Bowman, ``The repeval 2017
  shared task: Multi-genre natural language inference with sentence
  representations,'' \emph{arXiv preprint arXiv:1707.08172}, 2017.

\bibitem{White2017}
A.~S. White, P.~Rastogi, K.~Duh, and B.~Van~Durme, ``Inference is everything:
  Recasting semantic resources into a unified evaluation framework,'' in
  \emph{IJCNLP}, vol.~1, 2017, pp. 996--1005.

\bibitem{Pavlick2015}
E.~Pavlick, T.~Wolfe, P.~Rastogi, C.~Callison-Burch, M.~Dredze, and
  B.~Van~Durme, ``Framenet+: Fast paraphrastic tripling of framenet,'' in
  \emph{ACL}, vol.~2, 2015, pp. 408--413.

\bibitem{Rahman2012}
A.~Rahman and V.~Ng, ``Resolving complex cases of definite pronouns: the
  winograd schema challenge,'' in \emph{Joint EMNLP and CoNLL}, 2012.

\bibitem{Reisinger2015}
D.~Reisinger, R.~Rudinger, F.~Ferraro, C.~Harman, K.~Rawlins, and B.~Van~Durme,
  ``Semantic proto-roles,'' \emph{Transactions of the ACL}, vol.~3, pp.
  475--488, 2015.

\bibitem{Poliak2018b}
A.~Poliak, J.~Naradowsky, A.~Haldar, R.~Rudinger, and B.~Van~Durme,
  ``Hypothesis only baselines in natural language inference,'' \emph{arXiv
  preprint arXiv:1805.01042}, 2018.

\bibitem{Herzig2017}
J.~Herzig and J.~Berant, ``Neural semantic parsing over multiple
  knowledge-bases,'' \emph{arXiv preprint arXiv:1702.01569}, 2017.

\bibitem{Wang2015}
Y.~Wang, J.~Berant, and P.~Liang, ``Building a semantic parser overnight,'' in
  \emph{ACL and IJCNLP}, vol.~1, 2015, pp. 1332--1342.

\bibitem{Brunner2018}
G.~Brunner, Y.~Wang, R.~Wattenhofer, and M.~Weigelt, ``Natural language
  multitasking: Analyzing and improving syntactic saliency of hidden
  representations,'' \emph{arXiv preprint arXiv:1801.06024}, 2018.

\bibitem{Koehn2005}
P.~Koehn, ``Europarl: A parallel corpus for statistical machine translation,''
  in \emph{MT summit}, vol.~5, 2005, pp. 79--86.

\bibitem{Miller1995}
G.~A. Miller, ``Wordnet: a lexical database for english,'' \emph{Communications
  of the ACM}, vol.~38, no.~11, pp. 39--41, 1995.

\bibitem{Auer2007}
S.~Auer, C.~Bizer, G.~Kobilarov, J.~Lehmann, R.~Cyganiak, and Z.~Ives,
  ``Dbpedia: A nucleus for a web of open data,'' in \emph{The Semantic
  Web}.\hskip 1em plus 0.5em minus 0.4em\relax Springer, 2007, pp. 722--735.

\bibitem{Wang2017b}
Q.~Wang, Z.~Mao, B.~Wang, and L.~Guo, ``Knowledge graph embedding: A survey of
  approaches and applications,'' \emph{IEEE Transactions on Knowledge \& Data
  Engineering}, vol.~29, no.~12, pp. 2724--2743, 2017.

\bibitem{Hinton2012}
G.~Hinton, L.~Deng, D.~Yu, G.~E. Dahl, A.-r. Mohamed, N.~Jaitly, A.~Senior,
  V.~Vanhoucke, P.~Nguyen, T.~N. Sainath \emph{et~al.}, ``Deep neural networks
  for acoustic modeling in speech recognition: The shared views of four
  research groups,'' \emph{IEEE Signal Processing Magazine}, vol.~29, no.~6,
  pp. 82--97, 2012.

\bibitem{Graves2013}
A.~Graves, A.-r. Mohamed, and G.~Hinton, ``Speech recognition with deep
  recurrent neural networks,'' in \emph{IEEE International Conf on Acoustics,
  Speech and Signal Processing}, 2013, pp. 6645--6649.

\bibitem{Kenter2017}
T.~Kenter, A.~Borisov, C.~Van~Gysel, M.~Dehghani, M.~de~Rijke, and B.~Mitra,
  ``Neural networks for information retrieval,'' in \emph{Proceedings of the
  40th International ACM SIGIR Conference on Research and Development in
  Information Retrieval}.\hskip 1em plus 0.5em minus 0.4em\relax ACM, 2017, pp.
  1403--1406.

\bibitem{Huang2013}
P.-S. Huang, X.~He, J.~Gao, L.~Deng, A.~Acero, and L.~Heck, ``Learning deep
  structured semantic models for web search using clickthrough data,'' in
  \emph{ACM CIKM}, 2013, pp. 2333--2338.

\bibitem{Shen2014}
Y.~Shen, X.~He, J.~Gao, L.~Deng, and G.~Mesnil, ``Learning semantic
  representations using convolutional neural networks for web search,'' in
  \emph{Intnl Conf on World Wide Web}, 2014, pp. 373--374.

\bibitem{Lu2013}
Z.~Lu and H.~Li, ``A deep architecture for matching short texts,'' in
  \emph{Advances in neural information processing systems}, 2013.

\bibitem{Pang2016}
L.~Pang, Y.~Lan, J.~Guo, J.~Xu, S.~Wan, and X.~Cheng, ``Text matching as image
  recognition,'' in \emph{Thirtieth AAAI Conference on Artificial
  Intelligence}, 2016.

\bibitem{Guo2016b}
J.~Guo, Y.~Fan, Q.~Ai, and W.~B. Croft, ``A deep relevance matching model for
  ad-hoc retrieval,'' in \emph{Proceedings of the 25th ACM International on
  Conference on Information and Knowledge Management}.\hskip 1em plus 0.5em
  minus 0.4em\relax ACM, 2016, pp. 55--64.

\bibitem{Zamani2018}
H.~Zamani, M.~Dehghani, W.~B. Croft, E.~Learned-Miller, and J.~Kamps, ``From
  neural re-ranking to neural ranking: Learning a sparse representation for
  inverted indexing,'' in \emph{27th ACM International Conference on
  Information and Knowledge Management}.\hskip 1em plus 0.5em minus 0.4em\relax
  ACM, 2018.

\bibitem{MacAvaney2019}
S.~MacAvaney, A.~Yates, A.~Cohan, and N.~Goharian, ``Cedr: Contextualized
  embeddings for document ranking,'' \emph{CoRR}, 2019.

\bibitem{Cowie1996}
J.~Cowie and W.~Lehnert, ``Information extraction,'' \emph{CACM}, vol.~39,
  no.~1, pp. 80--91, 1996.

\bibitem{Hammerton2003}
J.~Hammerton, ``Named entity recognition with long short-term memory,'' in
  \emph{HLT-NAACL 2003-Volume 4}, 2003, pp. 172--175.

\bibitem{DosSantos2015}
C.~N.~d. Santos and V.~Guimaraes, ``Boosting named entity recognition with
  neural character embeddings,'' \emph{arXiv preprint arXiv:1505.05008}, 2015.

\bibitem{Santos2006}
D.~Santos, N.~Seco, N.~Cardoso, and R.~Vilela, ``Harem: An advanced ner
  evaluation contest for portuguese,'' in \emph{LREC, Genoa}, 2006.

\bibitem{Carreras2002}
X.~Carreras, L.~Marquez, and L.~Padr{\'o}, ``Named entity extraction using
  adaboost,'' in \emph{CoNLL}, 2002, pp. 1--4.

\bibitem{Chiu2015}
J.~Chiu and E.~Nichols, ``Named entity recognition with bidirectional
  lstm-cnns,'' \emph{arXiv preprint arXiv:1511.08308}, 2015.

\bibitem{Tjong2003}
E.~F. Tjong Kim~Sang and F.~De~Meulder, ``Introduction to the conll-2003 shared
  task: Language-independent named entity recognition,'' in
  \emph{HLT-NAACL-Volume 4}, 2003, pp. 142--147.

\bibitem{Hovy2006}
E.~Hovy, M.~Marcus, M.~Palmer, L.~Ramshaw, and R.~Weischedel, ``Ontonotes: the
  90\% solution,'' in \emph{Human Language Technology Conf, Companion Volume},
  2006, pp. 57--60.

\bibitem{Pradhan2013}
S.~Pradhan, A.~Moschitti, N.~Xue, H.~T. Ng, A.~Bj{\"o}rkelund, O.~Uryupina,
  Y.~Zhang, and Z.~Zhong, ``Towards robust linguistic analysis using
  ontonotes,'' in \emph{CoNLL}, 2013, pp. 143--152.

\bibitem{Lample2016}
G.~Lample, M.~Ballesteros, S.~Subramanian, K.~Kawakami, and C.~Dyer, ``Neural
  architectures for named entity recognition,'' \emph{arXiv preprint
  arXiv:1603.01360}, 2016.

\bibitem{Lafferty2001}
J.~Lafferty, A.~McCallum, and F.~C. Pereira, ``Conditional random fields:
  Probabilistic models for segmenting and labeling sequence data,'' 2001.

\bibitem{Akbik2018}
A.~Akbik, D.~Blythe, and R.~Vollgraf, ``Contextual string embeddings for
  sequence labeling,'' in \emph{COLING}, 2018, pp. 1638--1649.

\bibitem{Chen2015}
Y.~Chen, L.~Xu, K.~Liu, D.~Zeng, and J.~Zhao, ``Event extraction via dynamic
  multi-pooling convolutional neural networks,'' in \emph{ACL}, vol.~1, 2015,
  pp. 167--176.

\bibitem{Nguyen2016}
T.~H. Nguyen, K.~Cho, and R.~Grishman, ``Joint event extraction via recurrent
  neural networks,'' in \emph{Conf of the North American Chapter of ACL: Human
  Language Technologies}, 2016, pp. 300--309.

\bibitem{Liu2019c}
X.~Liu, H.~Huang, and Y.~Zhang, ``Open domain event extraction using neural
  latent variable models,'' \emph{arXiv preprint arXiv:1906.06947}, 2019.

\bibitem{Zheng2017}
S.~Zheng, Y.~Hao, D.~Lu, H.~Bao, J.~Xu, H.~Hao, and B.~Xu, ``Joint entity and
  relation extraction based on a hybrid neural network,''
  \emph{Neurocomputing}, vol. 257, pp. 59--66, 2017.

\bibitem{Sun2018}
M.~Sun, X.~Li, X.~Wang, M.~Fan, Y.~Feng, and P.~Li, ``Logician: A unified
  end-to-end neural approach for open-domain information extraction,'' in
  \emph{ACM Intnl Conf on Web Search and Data Mining}, 2018.

\bibitem{Tu2016}
Z.~Tu, Z.~Lu, Y.~Liu, X.~Liu, and H.~Li, ``Modeling coverage for neural machine
  translation,'' \emph{arXiv preprint arXiv:1601.04811}, 2016.

\bibitem{Lin2019}
C.~Lin, T.~Miller, D.~Dligach, S.~Bethard, and G.~Savova, ``A bert-based
  universal model for both within-and cross-sentence clinical temporal relation
  extraction,'' in \emph{Clinical NLP Workshop}, 2019, pp. 65--71.

\bibitem{Conneau2017}
A.~Conneau, H.~Schwenk, L.~Barrault, and Y.~Lecun, ``Very deep convolutional
  networks for text classification,'' in \emph{European ACL}, vol.~1, 2017, pp.
  1107--1116.

\bibitem{Jiang2018}
M.~Jiang, Y.~Liang, X.~Feng, X.~Fan, Z.~Pei, Y.~Xue, and R.~Guan, ``Text
  classification based on deep belief network and softmax regression,''
  \emph{Neural Computing and Applications}, vol.~29, no.~1, pp. 61--70, 2018.

\bibitem{Hinton2006}
G.~Hinton, S.~Osindero, and Y.-W. Teh, ``A fast learning algorithm for deep
  belief nets,'' \emph{Neural computation}, vol.~18, no.~7, 2006.

\bibitem{Sutton1998}
R.~S. Sutton and A.~G. Barto, \emph{Reinforcement learning: An
  introduction}.\hskip 1em plus 0.5em minus 0.4em\relax MIT Press Cambridge,
  1998, vol.~1, no.~1.

\bibitem{Smolensky1986}
P.~Smolensky, ``Information processing in dynamical systems: Foundations of
  harmony theory,'' Tech. Rep., 1986.

\bibitem{Fletcher2013}
R.~Fletcher, \emph{Practical methods of optimization}.\hskip 1em plus 0.5em
  minus 0.4em\relax Wiley \& Sons, 2013.

\bibitem{Adhikari2019}
A.~Adhikari, A.~Ram, R.~Tang, and J.~Lin, ``Docbert: Bert for document
  classification,'' \emph{arXiv preprint arXiv:1904.08398}, 2019.

\bibitem{Worsham2018}
J.~Worsham and J.~Kalita, ``Genre identification and the compositional effect
  of genre in literature,'' in \emph{COLING}, 2018, pp. 1963--1973.

\bibitem{Wei2018}
J.~Wei, Q.~Zhou, and Y.~Cai, ``Poet-based poetry generation: Controlling
  personal style with recurrent neural networks,'' in \emph{2018 International
  Conference on Computing, Networking and Communications (ICNC)}.\hskip 1em
  plus 0.5em minus 0.4em\relax IEEE, 2018, pp. 156--160.

\bibitem{Hopkins2017}
J.~Hopkins and D.~Kiela, ``Automatically generating rhythmic verse with neural
  networks,'' in \emph{Proceedings of the 55th Annual Meeting of the ACL
  (Volume 1: Long Papers)}, 2017, pp. 168--178.

\bibitem{Radford2019}
A.~Radford, J.~Wu, R.~Child, D.~Luan, D.~Amodei, and I.~Sutskever, ``Language
  models are unsupervised multitask learners,'' \emph{OpenAI Blog}, vol.~1,
  no.~8, 2019.

\bibitem{Bena2019}
B.~Bena and J.~Kalita, ``Introducing aspects of creativity in automatic poetry
  generation,'' in \emph{Intnl Conf on NLP}, 2019.

\bibitem{Tucker2019}
S.~Tucker and J.~Kalita, ``Genrating believable poetry in multiple languages
  using gpt-2,'' in \emph{Technical Report, University of Colorado, Colorado
  Springs}, 2019.

\bibitem{Yu2018}
Z.~Yu, J.~Tan, and X.~Wan, ``A neural approach to pun generation,'' in
  \emph{Proceedings of the 56th Annual Meeting of the Association for
  Computational Linguistics (Vol. 1: Long Papers)}, 2018, pp. 1650--1660.

\bibitem{Ren2017}
H.~Ren and Q.~Yang, ``Neural joke generation,'' \emph{Final Project Reports of
  Course CS224n}, 2017.

\bibitem{Chippada2018}
B.~Chippada and S.~Saha, ``Knowledge amalgam: Generating jokes and quotes
  together,'' \emph{arXiv preprint arXiv:1806.04387}, 2018.

\bibitem{Jain2017}
P.~Jain, P.~Agrawal, A.~Mishra, M.~Sukhwani, A.~Laha, and K.~Sankaranarayanan,
  ``Story generation from sequence of independent short descriptions,''
  \emph{arXiv preprint arXiv:1707.05501}, 2017.

\bibitem{Peng2018}
N.~Peng, M.~Ghazvininejad, J.~May, and K.~Knight, ``Towards controllable story
  generation,'' in \emph{Proceedings of the First Workshop on Storytelling},
  2018, pp. 43--49.

\bibitem{Martin2018}
L.~J. Martin, P.~Ammanabrolu, X.~Wang, W.~Hancock, S.~Singh, B.~Harrison, and
  M.~O. Riedl, ``Event representations for automated story generation with deep
  neural nets,'' in \emph{Thirty-Second AAAI Conference on Artificial
  Intelligence}, 2018.

\bibitem{Clark2018}
E.~Clark, Y.~Ji, and N.~A. Smith, ``Neural text generation in stories using
  entity representations as context,'' in \emph{Proceedings of the 2018
  Conference of the North American Chapter of the ACL: Human Language
  Technologies, Vol. 1 (Long Papers)}, 2018, pp. 2250--2260.

\bibitem{Xu2018}
J.~Xu, Y.~Zhang, Q.~Zeng, X.~Ren, X.~Cai, and X.~Sun, ``A skeleton-based model
  for promoting coherence among sentences in narrative story generation,''
  \emph{arXiv preprint arXiv:1808.06945}, 2018.

\bibitem{Drissi2018}
M.~Drissi, O.~Watkins, and J.~Kalita, ``Hierarchical text generation using an
  outline,'' \emph{Intl Conf on NLP}, 2018.

\bibitem{Huang2018}
Q.~Huang, Z.~Gan, A.~Celikyilmaz, D.~Wu, J.~Wang, and X.~He, ``Hierarchically
  structured reinforcement learning for topically coherent visual story
  generation,'' \emph{arXiv preprint arXiv:1805.08191}, 2018.

\bibitem{Fan2018}
A.~Fan, M.~Lewis, and Y.~Dauphin, ``Hierarchical neural story generation,''
  \emph{arXiv preprint arXiv:1805.04833}, 2018.

\bibitem{Li2015}
J.~Li, M.-T. Luong, and D.~Jurafsky, ``A hierarchical neural autoencoder for
  paragraphs and documents,'' \emph{arXiv preprint arXiv:1506.01057}, 2015.

\bibitem{Lin2017}
K.~Lin, D.~Li, X.~He, Z.~Zhang, and M.-T. Sun, ``Adversarial ranking for
  language generation,'' in \emph{Advances in Neural Information Processing
  Systems}, 2017, pp. 3155--3165.

\bibitem{Tambwekar2018}
P.~Tambwekar, M.~Dhuliawala, A.~Mehta, L.~J. Martin, B.~Harrison, and M.~O.
  Riedl, ``Controllable neural story generation via reward shaping,''
  \emph{arXiv preprint arXiv:1809.10736}, 2018.

\bibitem{Zhang2017}
Y.~Zhang, Z.~Gan, K.~Fan, Z.~Chen, R.~Henao, D.~Shen, and L.~Carin,
  ``Adversarial feature matching for text generation,'' in \emph{34th
  International Conference on Machine Learning-Vol 70}.\hskip 1em plus 0.5em
  minus 0.4em\relax JMLR, 2017.

\bibitem{Chen2018b}
L.~Chen, S.~Dai, C.~Tao, H.~Zhang, Z.~Gan, D.~Shen, Y.~Zhang, G.~Wang,
  R.~Zhang, and L.~Carin, ``Adversarial text generation via feature-mover's
  distance,'' in \emph{Advances in Neural Information Processing Systems},
  2018, pp. 4666--4677.

\bibitem{Guo2018}
J.~Guo, S.~Lu, H.~Cai, W.~Zhang, Y.~Yu, and J.~Wang, ``Long text generation via
  adversarial training with leaked information,'' in \emph{Thirty-Second AAAI
  Conference on Artificial Intelligence}, 2018.

\bibitem{Kingma2013}
D.~P. Kingma and M.~Welling, ``Auto-encoding variational bayes,'' \emph{arXiv
  preprint arXiv:1312.6114}, 2013.

\bibitem{Doersch2016}
C.~Doersch, ``Tutorial on variational autoencoders,'' \emph{arXiv preprint
  arXiv:1606.05908}, 2016.

\bibitem{Serban2017}
I.~V. Serban, A.~Sordoni, R.~Lowe, L.~Charlin, J.~Pineau, A.~Courville, and
  Y.~Bengio, ``A hierarchical latent variable encoder-decoder model for
  generating dialogues,'' in \emph{Thirty-First AAAI Conference on Artificial
  Intelligence}, 2017.

\bibitem{Hu2018}
Z.~Hu, Z.~Yang, X.~Liang, R.~Salakhutdinov, and E.~P. Xing, ``Toward controlled
  generation of text,'' in \emph{Proceedings of the 34th International
  Conference on Machine Learning-Volume 70}.\hskip 1em plus 0.5em minus
  0.4em\relax JMLR. org, 2017, pp. 1587--1596.

\bibitem{Wang2019}
W.~Wang, Z.~Gan, H.~Xu, R.~Zhang, G.~Wang, D.~Shen, C.~Chen, and L.~Carin,
  ``Topic-guided variational autoencoders for text generation,'' \emph{arXiv
  preprint arXiv:1903.07137}, 2019.

\bibitem{Holtzman2019}
A.~Holtzman, J.~Buys, M.~Forbes, and Y.~Choi, ``The curious case of neural text
  degeneration,'' \emph{arXiv preprint arXiv:1904.09751}, 2019.

\bibitem{Clark2019}
E.~Clark, A.~Celikyilmaz, and N.~A. Smith, ``Sentence mover, similarity:
  Automatic evaluation for multi-sentence texts,'' in \emph{Proceedings of the
  57th Annual Meeting of the Association for Computational Linguistics}, 2019,
  pp. 2748--2760.

\bibitem{Hashimoto2019}
T.~B. Hashimoto, H.~Zhang, and P.~Liang, ``Unifying human and statistical
  evaluation for natural language generation,'' \emph{arXiv preprint
  arXiv:1904.02792}, 2019.

\bibitem{Gatt2018}
A.~Gatt and E.~Krahmer, ``Survey of the state of the art in natural language
  generation: Core tasks, applications and evaluation,'' \emph{Journal of
  Artificial Intelligence Research}, vol.~61, pp. 65--170, 2018.

\bibitem{Hossain2019}
M.~Hossain, F.~Sohel, M.~F. Shiratuddin, and H.~Laga, ``A comprehensive survey
  of deep learning for image captioning,'' \emph{ACM Computing Surveys (CSUR)},
  vol.~51, no.~6, p. 118, 2019.

\bibitem{Liu2019a}
X.~Liu, Q.~Xu, and N.~Wang, ``A survey on deep neural network-based image
  captioning,'' \emph{The Visual Computer}, vol.~35, no.~3, 2019.

\bibitem{Krantz2018}
J.~Krantz and J.~Kalita, ``Abstractive summarization using attentive neural
  techniques,'' \emph{Intl Conf on NLP}, 2018.

\bibitem{Ranzato2015}
M.~Ranzato, S.~Chopra, M.~Auli, and W.~Zaremba, ``Sequence level training with
  recurrent neural networks,'' \emph{arXiv preprint arXiv:1511.06732}, 2015.

\bibitem{Gehring2017}
J.~Gehring, M.~Auli, D.~Grangier, D.~Yarats, and Y.~N. Dauphin, ``Convolutional
  sequence to sequence learning,'' \emph{arXiv preprint arXiv:1705.03122},
  2017.

\bibitem{Zhang2019}
H.~Zhang, Y.~Gong, Y.~Yan, N.~Duan, J.~Xu, J.~Wang, M.~Gong, and M.~Zhou,
  ``Pretraining-based natural language generation for text summarization,''
  \emph{arXiv preprint arXiv:1902.09243}, 2019.

\bibitem{Dong2015}
L.~Dong, F.~Wei, M.~Zhou, and K.~Xu, ``Question answering over freebase with
  multi-column convolutional neural networks,'' in \emph{ACL and International
  Joint Conf on NLP}, vol.~1, 2015, pp. 260--269.

\bibitem{Santoro2017}
A.~Santoro, D.~Raposo, D.~G. Barrett, M.~Malinowski, R.~Pascanu, P.~Battaglia,
  and T.~Lillicrap, ``A simple neural network module for relational
  reasoning,'' in \emph{NIPS}, 2017, pp. 4974--4983.

\bibitem{Raposo2017}
D.~Raposo, A.~Santoro, D.~Barrett, R.~Pascanu, T.~Lillicrap, and P.~Battaglia,
  ``Discovering objects and their relations from entangled scene
  representations,'' \emph{arXiv preprint arXiv:1702.05068}, 2017.

\bibitem{Yang2019}
W.~Yang, Y.~Xie, A.~Lin, X.~Li, L.~Tan, K.~Xiong, M.~Li, and J.~Lin,
  ``End-to-end open-domain question answering with bertserini,'' \emph{arXiv
  preprint arXiv:1902.01718}, 2019.

\bibitem{Schwenk2012}
H.~Schwenk, ``Continuous space translation models for phrase-based statistical
  machine translation,'' \emph{COLING}, pp. 1071--1080, 2012.

\bibitem{Deselaers2009}
T.~Deselaers, S.~Hasan, O.~Bender, and H.~Ney, ``A deep learning approach to
  machine transliteration,'' in \emph{Workshop on Statistical Machine
  Translation}.\hskip 1em plus 0.5em minus 0.4em\relax ACL, 2009, pp. 233--241.

\bibitem{Kalchbrenner2013}
N.~Kalchbrenner and P.~Blunsom, ``Recurrent continuous translation models,'' in
  \emph{EMNLP}, 2013, pp. 1700--1709.

\bibitem{Collobert2008}
R.~Collobert and J.~Weston, ``A unified architecture for natural language
  processing: Deep neural networks with multitask learning,'' in \emph{ICML},
  2008, pp. 160--167.

\bibitem{Cho2014b}
K.~Cho, B.~Van~Merri{\"e}nboer, C.~Gulcehre, D.~Bahdanau, F.~Bougares,
  H.~Schwenk, and Y.~Bengio, ``Learning phrase representations using rnn
  encoder-decoder for statistical machine translation,'' \emph{arXiv preprint
  arXiv:1406.1078}, 2014.

\bibitem{Sutskever2014}
I.~Sutskever, O.~Vinyals, and Q.~V. Le, ``Sequence to sequence learning with
  neural networks,'' in \emph{NIPS}, 2014, pp. 3104--3112.

\bibitem{Wu2016b}
Y.~Wu, M.~Schuster, Z.~Chen, Q.~V. Le, M.~Norouzi, W.~Macherey, M.~Krikun,
  Y.~Cao, Q.~Gao, K.~Macherey \emph{et~al.}, ``Google's neural machine
  translation system: Bridging the gap between human and machine translation,''
  \emph{arXiv preprint arXiv:1609.08144}, 2016.

\bibitem{Britz2017}
D.~Britz, A.~Goldie, T.~Luong, and Q.~Le, ``Massive exploration of neural
  machine translation architectures,'' \emph{arXiv preprint arXiv:1703.03906},
  2017.

\bibitem{Sennrich2017}
R.~Sennrich, O.~Firat, K.~Cho, A.~Birch, B.~Haddow, J.~Hitschler,
  M.~Junczys-Dowmunt, S.~L{\"a}ubli, A.~V.~M. Barone, J.~Mokry \emph{et~al.},
  ``Nematus: a toolkit for neural machine translation,'' \emph{arXiv preprint
  arXiv:1703.04357}, 2017.

\bibitem{Klein2017}
G.~Klein, Y.~Kim, Y.~Deng, J.~Senellart, and A.~M. Rush, ``Opennmt: Open-source
  toolkit for neural machine translation,'' \emph{arXiv preprint
  arXiv:1701.02810}, 2017.

\bibitem{Sennrich2016}
R.~Sennrich and B.~Haddow, ``Linguistic input features improve neural machine
  translation,'' \emph{arXiv preprint arXiv:1606.02892}, 2016.

\bibitem{Ahmed2017}
K.~Ahmed, N.~S. Keskar, and R.~Socher, ``Weighted transformer network for
  machine translation,'' \emph{arXiv preprint arXiv:1711.02132}, 2017.

\bibitem{Hochreiter2001}
S.~Hochreiter, Y.~Bengio, P.~Frasconi, J.~Schmidhuber \emph{et~al.}, ``Gradient
  flow in recurrent nets: the difficulty of learning long-term dependencies,''
  2001.

\bibitem{Cettolo2014}
M.~Cettolo, J.~Niehues, S.~St{\"u}ker, L.~Bentivogli, and M.~Federico, ``Report
  on the 11th iwslt evaluation campaign, iwslt 2014,'' in \emph{International
  Workshop on Spoken Language Translation, Hanoi}, 2014.

\bibitem{Medina2018}
J.~R. Medina and J.~Kalita, ``Parallel attention mechanisms in neural machine
  translation,'' \emph{arXiv preprint arXiv:1810.12427}, 2018.

\bibitem{Papineni2002}
K.~Papineni, S.~Roukos, T.~Ward, and W.-J. Zhu, ``Bleu: a method for automatic
  evaluation of machine translation,'' in \emph{ACL}, 2002, pp. 311--318.

\bibitem{Ghazvininejad2019}
M.~Ghazvininejad, O.~Levy, Y.~Liu, and L.~Zettlemoyer, ``Constant-time machine
  translation with conditional masked language models,'' \emph{arXiv preprint
  arXiv:1904.09324}, 2019.

\bibitem{Lample2019}
G.~Lample and A.~Conneau, ``Cross-lingual language model pretraining,''
  \emph{arXiv preprint arXiv:1901.07291}, 2019.

\bibitem{Chen2018}
M.~X. Chen, O.~Firat, A.~Bapna, M.~Johnson, W.~Macherey, G.~Foster, L.~Jones,
  N.~Parmar, M.~Schuster, Z.~Chen \emph{et~al.}, ``The best of both worlds:
  Combining recent advances in neural machine translation,'' \emph{arXiv
  preprint arXiv:1804.09849}, 2018.

\bibitem{Denkowski2017}
M.~Denkowski and G.~Neubig, ``Stronger baselines for trustable results in
  neural machine translation,'' \emph{arXiv preprint arXiv:1706.09733}, 2017.

\bibitem{Koehn2017}
P.~Koehn and R.~Knowles, ``Six challenges for neural machine translation,''
  \emph{arXiv preprint arXiv:1706.03872}, 2017.

\bibitem{Kuang2018}
S.~Kuang, D.~Xiong, W.~Luo, and G.~Zhou, ``Modeling coherence for neural
  machine translation with dynamic and topic caches,'' in \emph{COLING}, 2018,
  pp. 596--606.

\bibitem{Luong2014}
M.-T. Luong, I.~Sutskever, Q.~V. Le, O.~Vinyals, and W.~Zaremba, ``Addressing
  the rare word problem in neural machine translation,'' \emph{arXiv preprint
  arXiv:1410.8206}, 2014.

\bibitem{Sennrich2015}
R.~Sennrich, B.~Haddow, and A.~Birch, ``Neural machine translation of rare
  words with subword units,'' \emph{arXiv preprint arXiv:1508.07909}, 2015.

\bibitem{Mager2018}
M.~Mager, E.~Mager, A.~Medina-Urrea, I.~Meza, and K.~Kann, ``Lost in
  translation: Analysis of information loss during machine translation between
  polysynthetic and fusional languages,'' \emph{arXiv preprint
  arXiv:1807.00286}, 2018.

\bibitem{Ott2018}
M.~Ott, M.~Auli, D.~Granger, and M.~Ranzato, ``Analyzing uncertainty in neural
  machine translation,'' \emph{arXiv preprint arXiv:1803.00047}, 2018.

\bibitem{Wang2018b}
W.~Wang, T.~Watanabe, M.~Hughes, T.~Nakagawa, and C.~Chelba, ``Denoising neural
  machine translation training with trusted data and online data selection,''
  \emph{arXiv preprint arXiv:1809.00068}, 2018.

\bibitem{Johnson2016}
M.~Johnson, M.~Schuster, Q.~V. Le, M.~Krikun, Y.~Wu, Z.~Chen, N.~Thorat,
  F.~Vi{\'e}gas, M.~Wattenberg, G.~Corrado \emph{et~al.}, ``Google's
  multilingual neural machine translation system: enabling zero-shot
  translation,'' \emph{arXiv preprint arXiv:1611.04558}, 2016.

\bibitem{Jurafsky2017}
D.~Jurafsky and J.~Martin, \emph{Speech \& language processing (3rd. Edition
  Draft)}.\hskip 1em plus 0.5em minus 0.4em\relax Pearson Education, 2017.

\bibitem{Zheng2018}
L.~Zheng, H.~Wang, and S.~Gao, ``Sentimental feature selection for sentiment
  analysis of chinese online reviews,'' \emph{Intnl J. of Machine Learning and
  Cybernetics}, vol.~9, no.~1, pp. 75--84, 2018.

\bibitem{Etter2018}
M.~Etter, E.~Colleoni, L.~Illia, K.~Meggiorin, and A.~D'Eugenio, ``Measuring
  organizational legitimacy in social media: Assessing citizens' judgments with
  sentiment analysis,'' \emph{Business \& Society}, vol.~57, no.~1, pp. 60--97,
  2018.

\bibitem{Cliche2017}
M.~Cliche, ``Bb\_twtr at semeval-2017 task 4: Twitter sentiment analysis with
  cnns and lstms,'' \emph{arXiv preprint arXiv:1704.06125}, 2017.

\bibitem{Bobrow1964}
D.~G. Bobrow, ``Natural language input for a computer problem solving system,''
  1964.

\bibitem{Weizenbaum1966}
J.~Weizenbaum, ``Eliza, a computer program for the study of natural language
  communication between man and machine,'' \emph{CACM}, vol.~9, no.~1, pp.
  36--45, 1966.

\bibitem{Winograd1971}
T.~Winograd, ``Procedures as a representation for data in a computer program
  for understanding natural language,'' MIT, Tech. Rep., 1971.

\bibitem{Wang2018c}
A.~Wang, A.~Singh, J.~Michael, F.~Hill, O.~Levy, and S.~R. Bowman, ``Glue: A
  multi-task benchmark and analysis platform for natural language
  understanding,'' \emph{arXiv preprint arXiv:1804.07461}, 2018.

\bibitem{Schuman2017}
C.~D. Schuman, T.~E. Potok, R.~M. Patton, J.~D. Birdwell, M.~E. Dean, G.~S.
  Rose, and J.~S. Plank, ``A survey of neuromorphic computing and neural
  networks in hardware,'' \emph{arXiv preprint arXiv:1705.06963}, 2017.

\bibitem{Hennessy2017}
J.~Hennessy and D.~Patterson, \emph{Computer architecture: a quantitative
  approach}.\hskip 1em plus 0.5em minus 0.4em\relax Elsevier, 2017.

\bibitem{Monroe2014}
D.~Monroe, ``Neuromorphic computing gets ready for the (really) big time,''
  \emph{CACM}, vol.~57, no.~6, pp. 13--15, 2014.

\end{thebibliography}

\end{document}